\documentclass{article} 
\usepackage{iclr2026_conference,times}

\usepackage{amsmath,amsfonts,bm}

\def\eqref#1{equation~\ref{#1}}

\def\1{\bm{1}}

\DeclareMathAlphabet{\mathsfit}{\encodingdefault}{\sfdefault}{m}{sl}
\SetMathAlphabet{\mathsfit}{bold}{\encodingdefault}{\sfdefault}{bx}{n}

\usepackage{hyperref}
\usepackage{url}

\usepackage{enumitem}
\usepackage{amsmath}
\usepackage{multirow}
\usepackage{adjustbox}
\usepackage{array}
\usepackage{booktabs} 
\usepackage{tabularx}
\usepackage{graphicx}
\usepackage{subcaption}

\usepackage{algorithm}        
\usepackage{algpseudocode}    

\usepackage{subcaption}

\usepackage{amssymb}
\usepackage{pifont}

\usepackage[most]{tcolorbox}

\definecolor{codegreen}{rgb}{0,0.6,0}
\definecolor{codegray}{rgb}{0.5,0.5,0.5}
\definecolor{codepurple}{rgb}{0.58,0,0.82}
\definecolor{backcolour}{rgb}{0.95,0.95,0.92}
\usepackage{listings}
\lstdefinestyle{mystyle}{
  commentstyle=\color{codegreen},
  keywordstyle=\color{magenta},
  numberstyle=\small\color{codegray},
  stringstyle=\color{codepurple},
  basicstyle=\small,
  breakatwhitespace=false,         
  breaklines=true,                 
  captionpos=b,                    
  keepspaces=false,                                 
  showspaces=false,                
  showstringspaces=false,
  showtabs=false,                  
  tabsize=2
}

\newcommand\blfootnote[1]{%
\begingroup
\renewcommand\thefootnote{}\footnote{#1}%
\addtocounter{footnote}{-1}%
\endgroup
}

\title{WebGen-Agent: Enhancing Interactive Website Generation with Multi-Level Feedback and Step-Level Reinforcement Learning}

\author{%
  Zimu Lu$^{1*}$, Houxing Ren$^{1*}$, Yunqiao Yang$^{1}$, Ke Wang$^{1}$, Zhuofan Zong$^{1}$ \\ {\bf Junting Pan$^{1}$}, {\bf Mingjie Zhan}$^{1\dagger}$, {\bf Hongsheng Li}$^{1,2\dagger}$\\
  \textsuperscript{1}Multimedia Laboratory (MMLab), The Chinese University of Hong Kong, \textsuperscript{2}Ace Robotics\\
 \small{
   \href{mailto:email@domain}{luzimu@link.cuhk.edu.hk} $\quad$
   \href{mailto:email@domain}{zhanmingjie@sensetime.com} $\quad$
   \href{mailto:email@domain}{hsli@ee.cuhk.edu.hk}
 }
}

\begin{document}

\maketitle

\begin{abstract}
Agent systems powered by large language models (LLMs) have demonstrated impressive performance on repository-level code-generation tasks. However, for tasks such as website codebase generation, which depend heavily on visual effects and user-interaction feedback, current code agents rely only on simple code execution for feedback and verification. This approach fails to capture the actual quality of the generated code.
In this paper, we propose \textit{WebGen-Agent}, a novel website-generation agent that leverages comprehensive and multi-level visual feedback to iteratively generate and refine the website codebase. Detailed and expressive text descriptions and suggestions regarding the screenshots and GUI-agent testing of the websites are generated by a visual language model (VLM), together with scores that quantify their quality. The screenshot and GUI-agent scores are further integrated with a backtracking and select-best mechanism, enhancing the performance of the agent.
Utilizing the accurate visual scores inherent in the WebGen-Agent workflow, we further introduce \textit{Step-GRPO with Screenshot and GUI-agent Feedback} to improve the ability of LLMs to act as the reasoning engine of WebGen-Agent. By using the screenshot and GUI-agent scores at each step as the reward in Step-GRPO, we provide a dense and reliable process supervision signal, which effectively improves the model's website-generation ability.
On the WebGen-Bench dataset, WebGen-Agent increases the accuracy of Claude-3.5-Sonnet from 26.4\% to 51.9\% and its appearance score from 3.0 to 3.9, outperforming the previous state-of-the-art agent system. Additionally, our Step-GRPO training approach increases the accuracy of Qwen2.5-Coder-7B-Instruct from 38.9\% to 45.4\% and raises the appearance score from 3.4 to 3.7. We release the WebGen-Agent workflow code, along with the training code, data, and model weights at~\url{https://github.com/mnluzimu/WebGen-Agent}.
\end{abstract}

\blfootnote{$^*$Equal contribution\quad $^\dagger$Corresponding author}

\section{Introduction}

Recent studies on code agents have shown great advancements in repository-level code-generation tasks, such as fixing GitHub issues~\citep{NEURIPS2024_5a7c9475} and implementing new features~\citep{miserendino2025swelancerfrontierllmsearn}. However, for tasks like website code generation, which depend heavily on visual aesthetics and the fluency of user interactions, current code-agent systems fail to fully capture the actual quality of the generated codebase, because they mostly rely on simple code-execution feedback. This limitation can lead to various rendering and functional problems in the generated web applications, such as misaligned components, disharmonious coloring, unresponsive buttons, and broken links.

To enable the code agent to effectively handle such tasks, we introduce \textbf{WebGen-Agent}, a code-generation system that generates websites from natural-language instructions that specify appearance and functional requirements, thus offering a highly automated website-development process. To ensure that the generated websites meet both functional requirements and aesthetic standards, we leverage both execution feedback and visual feedback to refine the project. Specifically, we leverage a visual language model (VLM) to assess the visual appeal and aesthetic quality of the current website, and a graphical user interface (GUI) agent to evaluate the correctness and intended functionality of the website’s codebase, thereby gathering accurate information and providing targeted suggestions. By iteratively applying this feedback and editing the codebase, WebGen-Agent builds websites with appealing designs and smooth interactive functionality.

As shown in Fig.~\ref{fig:webgen_agent_workflow}, WebGen-Agent adopts an iterative, multi-step paradigm in which each step consists of three actions: code generation, code execution, and feedback gathering. The agent begins each step by creating and editing files in the codebase in a manner similar to Bolt.diy~\citep{stackblitzlabs2024bolt}. During code execution, dependencies are installed, and the website service is started. If execution emits errors, the errors are returned to the agent, which starts the next step to fix them. If five consecutive erroneous steps occur, the agent backtracks to a previous non-erroneous step.

In the feedback-gathering process, a screenshot of the landing page of the website is captured first. A VLM then provides a description and an appearance score based on the screenshot. If the screenshot has room for improvement, the model supplies suggestions, which are implemented in the subsequent step to explicitly refine the website’s visual aesthetics. Otherwise, a GUI-agent session is initiated to explore the website, which evaluates the functional requirements and generates corresponding feedback. If the testing is successful, the task is complete; otherwise, suggestions for fixing the website are generated, and the agent can edit the codebase in the next step. At the end of the task trajectory, the best step is selected on the basis of the screenshot and GUI-agent scores, and the codebase is restored to the state of that step. Based on the pipeline, various models achieve better performance on WebGen-Bench~\citep{lu2025webgenbenchevaluatingllmsgenerating}, consistently outperforming other code agents. Remarkably, Claude-3.5-Sonnet improves its accuracy from 26.4\% to 51.9\% and its appearance score from 3.0 to 3.9, outperforming Bolt.diy.

To equip code agents with enhanced reasoning abilities, we further propose \textbf{Step-GRPO with Screenshot and GUI-agent Feedback}. Given an instruction, multiple WebGen-Agent trajectories are generated. Each step in an agent trajectory is accompanied by a screenshot score and a GUI-agent testing score, and an accurate and reliable step-level reward can be computed by summing these two scores. This dual supervision of website appearance and functionality effectively optimizes the model to generate high-quality website codebases, providing stepwise, process-level guidance for the agent trajectory. Training a Qwen2.5-Coder-7B-Instruct model with this Step-GRPO approach increases the accuracy from 38.9\% to 45.4\% and raises the appearance score from 3.4 to 3.7 on WebGen-Bench, greatly improving both the functionality and the appearance of the generated websites. We name the trained family of models \textbf{WebGenAgent-LM}.

Our contributions include:

\begin{itemize}[leftmargin=0.4cm]
    \item We propose WebGen-Agent, a code-agent system that leverages screenshots and GUI-agent testing to provide feedback signals and iteratively improve the quality of generated websites.
    \item We introduce Step-GRPO with Screenshot and GUI-agent Feedback, which uses screenshots and GUI-agent scores as step-level supervision in the GRPO training process, significantly improving the performance of smaller open-source models.
    \item Extensive experiments demonstrate the effectiveness of the proposed system. The system increases the accuracy of Claude-3.5-Sonnet from 26.4\% to 51.9\% and its appearance score from 3.0 to 3.9, outperforming Bolt.diy. Our training approach also increases the accuracy of Qwen2.5-Coder-7B-Instruct from 38.9\% to 45.4\% and raises the appearance score from 3.4 to 3.7.
\end{itemize}

\section{Method}

In this section, we first introduce WebGen-Agent, a novel website generation system that leverages screenshots and GUI-agent testing as reliable feedback to iteratively refine both the appearance and functionalities of the generated website with a coding LLM. Building on the reliable visual scores produced by WebGen-Agent, we then propose Step-GRPO with Screenshot and GUI-agent Feedback, a method that uses these scores to provide process supervision during GRPO training. This system significantly enhances language models' ability to generate high-quality websites.


\subsection{WebGen-Agent Workflow}
\label{sec:webgen-agent-workflow}

\begin{figure*}[t]
    \centering
    \includegraphics[width=1.0\textwidth,page=1]{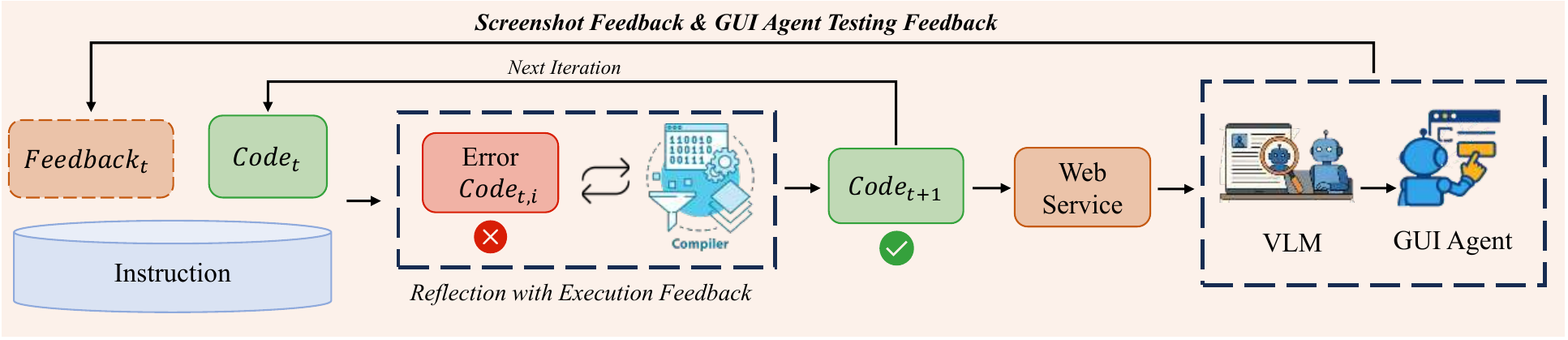}
    \caption{Iterative website generation with screenshot- and GUI-agent-based feedback. A backtracking and best-step-selection mechanism is applied on the basis of the screenshot and GUI-agent testing scores.}
    \label{fig:webgen_agent_workflow}
\end{figure*}

The WebGen-Agent workflow consists of multiple steps, with each step including code generation, code execution, and feedback gathering. As shown in Fig.~\ref{fig:webgen_agent_workflow}, the agent trajectory starts from a website generation instruction ($\mathcal{I}$), denoted as $\mathcal{T} = [\mathcal{I}]$, and an empty codebase $\mathcal{C}_0$. The instruction $\mathcal{I}$ is created by concatenating a system prompt similar to that of Bolt.diy~\citep{stackblitzlabs2024bolt} with the user-provided website-generation request. A coding LLM acting as the engine of the agent generates code $\Delta\mathcal{C}_1$ to edit the codebase, resulting in $\mathcal{C}_1$. Then, the dependencies of the codebase are installed, and the website service is started. The code execution output is denoted as $\mathcal{O}_1$, which contains both stdout and stderr. If the dependency installation or service initialization fails, the output message $\mathcal{O}_1$ is returned to the agent as feedback, so that the agent can fix the error in the next step.
If no error occurs, a screenshot of the website is captured and presented to a VLM (e.g. Qwen2.5-VL-32B;~\cite{bai2025qwen25vltechnicalreport}), which is requested to provide a description of the screenshot and, if needed, suggestions to improve the website's appearance. The prompt for acquiring screenshot feedback is provided in Fig.~\ref{fig:prompt_screenshot_desc} of Appendix~\ref{sec:webgen_agent_prompts}. A score of the website appearance based on the screenshot is also generated and, together with the description and suggestions, composes the screenshot feedback. The feedback can be denoted as:

\begin{equation}
\mathcal{F}_\text{shot} = \bigl\langle\textit{Description},\textit{Score}_{\text{shot}},\textit{Suggestions}_{\text{shot}}\bigr\rangle
\end{equation}

$\mathcal{F}_\text{shot}$ is used to reflects the integrity and aesthetics of the website's appearance. Here, a separate VLM is used besides the coding LLM to make the system more cost-effective, as we observe that a relatively small open-source VLM is sufficient for the task, while the code generation requires an LLM with strong coding abilities. We use Qwen2.5-VL-32B-Instruct as the VLM in our experiments unless stated otherwise. The code execution and screenshot feedback are appended to the agent trajectory, resulting in $\mathcal{T} = [\mathcal{I}, \Delta\mathcal{C}_1, \mathcal{O}_1, \mathcal{F}_\text{shot,1}]$. Then, the agent judges whether the website's appearance is satisfactory based on the trajectory. If it is unsatisfactory, the agent continues to generate code $\Delta\mathcal{C}_2$ to improve the website's appearance. Otherwise, the agent initiates a GUI-agent testing session,  generating an instruction for the GUI-agent to explore various website functionalities specified in the instruction $\mathcal{I}$, resulting in a GUI-agent testing trajectory. The prompt used to generate the GUI-agent instructions is shown in Fig.~\ref{fig:gui_agent_trigger_prompt} of Appendix~\ref{sec:webgen_agent_prompts}. It instructs the model to produce a GUI-agent instruction that comprehensively checks all website-development requirements and includes a one-shot example. As shown in Tab.~\ref{tab:gui_instruction_scores} of Appendix~\ref{sec:gui_instruction_comprehensiveness}, a manual inspection indicates that 98.3\% of the sampled instructions achieve high coverage of the requirements. Based on the GUI-agent testing result, the LLM acting as the engine of the agent judges whether the testing is successful and provides a score, denoted as $\textit{Score}_{\text{gui}}$. The prompt for acquiring the GUI-agent testing feedback is provided in Fig.~\ref{fig:gui_agent_testing_prompt} of Appendix~\ref{sec:webgen_agent_prompts}. If the testing result is unsatisfactory, suggestions are also made to improve the functionality. Thus, the GUI-agent testing feedback can be denoted as:

\begin{equation}
\mathcal{F}_\text{gui} = \bigl\langle\textit{Score}_{\text{gui}},\textit{Suggestions}_{\text{gui}}\bigr\rangle
\end{equation}
$\mathcal{F}_\text{gui}$ is also appended to the trajectory, resulting in $\mathcal{T} = [\mathcal{I}, \Delta\mathcal{C}_1, \mathcal{O}_1, \mathcal{F}_\text{shot,1}, \mathcal{F}_\text{gui,1}]=[\mathcal{I}, \Delta\mathcal{C}_1, \mathcal{O}_1, \mathcal{F}_1]$. Here, $\mathcal{F}_1$ denotes $[\mathcal{F}_\text{shot,1}, \mathcal{F}_\text{gui,1}]$. In this way, WebGen-Agent continues to improve the appearance and functionality of the website, resulting in a trajectory $\mathcal{T}$, denoted as $\mathcal{T} = [\mathcal{I}, \Delta\mathcal{C}_1, \mathcal{O}_1, \mathcal{F}_1, \Delta\mathcal{C}_2, \mathcal{O}_2, \mathcal{F}_2, \dots, \Delta\mathcal{C}_K, \mathcal{O}_K, \mathcal{F}_K]$.

The process ends when the website passes the GUI-agent testing, or the maximum iteration number is reached. During the iterations, at step $i \in \{1, 2, \dots\}$, the codebase state $\mathcal{C}_i$, the edit $\Delta\mathcal{C}_i$, together with the $\textit{Score}_{\text{shot},i}$ and $\textit{Score}_{\text{gui},i}$, are stored in a memory list. If five consecutive steps contain code execution errors, a backtracking mechanism is triggered, and the agent trajectory and the codebase are returned to the state at the best previous step. The best previous step is selected by first choosing the steps with the highest $\textit{Score}_{\text{gui}}$, and then among these steps, the ones with the highest $\textit{Score}_{\text{shot}}$ are chosen. If there are still more than one chosen step, then the latest one among them is selected. Considering that later code edits might not always improve the previous codebase, at the end of the agent workflow, the best step among all the steps is selected in the same way as mentioned above. A more detailed algorithmic presentation can be found in Appendix~\ref{sec:webgen_agent_algorithm} and example trajectories are presented in Appendix~\ref{sec:trajectory_examples}.

\subsection{Step-GRPO with Screenshot and GUI-agent Feedback}
\label{sec:step-grpo}

\begin{figure*}[t]
    \centering
    \includegraphics[width=1.0\textwidth,page=2]{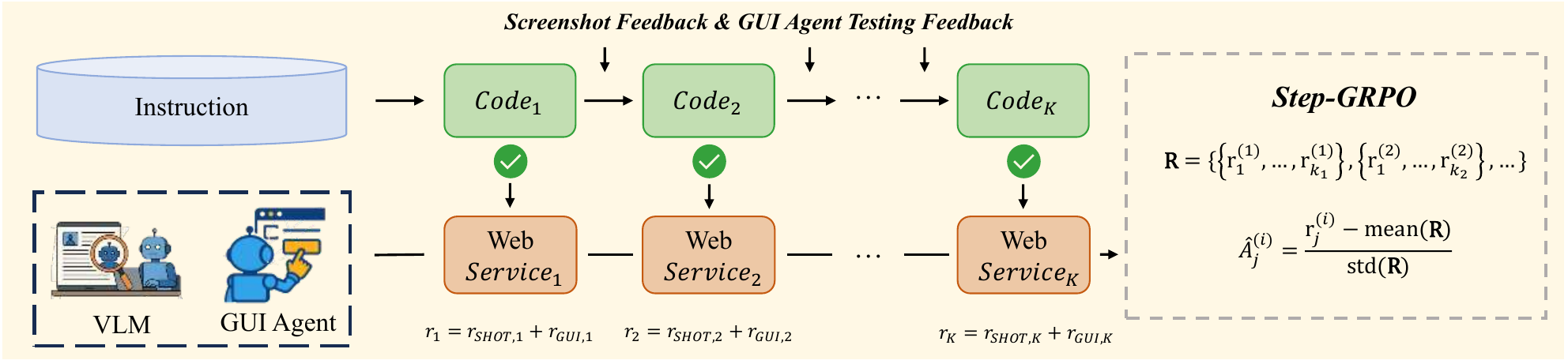}
    \caption{Step-GRPO with Screenshot and GUI-agent Feedback. Multiple WebGen-Agent trajectories are produced, and the reward for each step is computed by summing the screenshot score and the GUI-agent score.}
    \label{fig:webgen_agent_step_grpo}
\end{figure*}

While using strong proprietary models as the engine LLM in WebGen-Agent can produce high performance, the agent workflow would be more cost-efficient if smaller open-source models of 7B-8B parameters can be used instead. However, current small open-source language models still lag behind proprietary models in website code generation. Therefore, we introduce Step-GRPO with Screenshot and GUI-agent Feedback, leveraging the $\textit{Score}_{\text{shot}}$ and $\textit{Score}_{\text{gui}}$ inherently produced in the WebGen-Agent workflow to train them with step-level process supervision in GRPO training. 

Before the GRPO-based training, we first perform a light supervised fine-tuning (SFT) using approximately 700 WebGen-Agent trajectories generated by DeepSeek-V3, training for one epoch to serve as a warm start. Then, Step-GRPO is performed on the fine-tuned model. The Step-GRPO training objective can be written as:
\begin{equation}
\resizebox{0.94\linewidth}{!}{ 
$
\begin{aligned}
    \mathcal{J}_{GRPO}(\theta) &= \mathbb{E}_{[q \sim P(Q), \{o_i\}_{i=1}^G \sim \pi_{\theta_{old}}(O|q)]} \\
    &\quad \frac{1}{G}\sum_{i=1}^G\frac{1}{|o_i|} \sum_{t=1}^{|o_i|} \left\{ \min \left[ \frac{\pi_\theta(o_{i,t} | q, o_{i,<t})}{\pi_{\theta_{old}}(o_{i,t} | q, o_{i,<t})} \hat{A}_{i,t}, \text{clip} \left( \frac{\pi_\theta(o_{i,t} | q, o_{i,<t})}{\pi_{\theta_{old}}(o_{i,t} | q, o_{i,<t})}, 1 - \epsilon, 1 + \epsilon \right) \hat{A}_{i,t} \right] \right\},
\end{aligned}
$
}
\label{eq:GRPO-obj}
\end{equation}
Here, $q$ denotes the website generation instruction, and $\{o_i\}_{i=1}^G$ denotes the group of trajectories generated from the instruction $q$. We remove the KL loss to encourage the model to more freely adapt its behavior to the reward signals~\citep{qian2025toolrlrewardtoollearning}. $o_i$ can be denoted as $[\Delta\mathcal{C}_1, \mathcal{O}_1, \mathcal{F}_1,\dots, \Delta\mathcal{C}_{K_i}, \mathcal{O}_{K_i}, \mathcal{F}_{K_i}]$. Different from the naive GRPO, which sets the advantages on all tokens in a trajectory to the same value, the Step-GRPO sets advantages on tokens in different steps to different values. In our work, the GRPO loss is only applied to the model outputs ${\Delta\mathcal{C}_1,\Delta\mathcal{C}_2,\dots, \Delta\mathcal{C}_K}$. We denote the reward of all tokens in the $j$-th step of $o^{(i)}$ as $r_j^{(i)}$, which is computed by summing the $\textit{Score}_{\text{shot}}$ and $\textit{Score}_{\text{gui}}$ of that step, generated in the WebGen-Agent workflow:
\begin{equation}
r_j^{(i)} = \textit{Score}_{\text{shot}, j}^{(i)} + \textit{Score}_{\text{gui}, j}^{(i)}
\end{equation}
The rewards for all steps in the trajectories sampled from $q$ can be written as
$\mathbf{R}=\{ \{r_1^{(1)},\cdots,r_{K_1}^{(1)}\}, \dots,  \{r_1^{(G)},\cdots,r_{K_G}^{(G)}\} \}$.
The advantage for step $j$ of the $i$-th trajectory is computed by standardizing its immediate reward:
$\displaystyle \hat{A}_{j}^{(i)} = \frac{r_j^{(i)} - {\rm mean(\mathbf{R})}}{{\rm std(\mathbf{R})}}$. $\hat{A}_{j}^{(i)}$ denotes the advantage of $o^{(i)}$ at the $j$-th step. We do not accumulate normalized rewards from future steps as in~\cite{shao2024deepseekmathpushinglimitsmathematical}, because in the website-generation task $\textit{Score}_{\text{shot}}$ and $\textit{Score}_{\text{gui}}$ directly reflect the quality of the website at the current step, which is more appropriate for representing the desirability of the current code. The Step-GRPO training process is illustrated in Fig.~\ref{fig:webgen_agent_step_grpo}. This Step-GPPO method, with screenshot and GUI-agent feedback, incorporates accurate step-level supervision and effectively helps the model learn to generate websites with an appealing appearance and smooth functionality.

\section{Experiments}

In this section, we first present the performance of WebGen-Agent on WebGen-Bench using a variety of proprietary and open-source LLMs, as well as models trained using Step-GRPO with Screenshot and GUI-agent Feedback. Then, we conduct comprehensive ablation studies on the design choices in the WebGen-Agent workflow and the Step-GRPO training process.

\subsection{Main Results}
\label{sec:main-results}

\begin{table}[t]\fontsize{9}{10}\selectfont
\centering
\caption{The performance of WebGen-Agent with various proprietary and open-source models on WebGen-Bench~\citep{lu2025webgenbenchevaluatingllmsgenerating}, compared with other code agent systems. The highest Accuracy and Appearance Score are highlighted in \textbf{bold}.}
\begin{tabularx}{\textwidth}
  {>{\raggedright\arraybackslash\hsize=2.5\hsize}X
   >{\centering\arraybackslash\hsize=0.4\hsize}X
   >{\centering\arraybackslash\hsize=0.6\hsize}X
   >{\centering\arraybackslash\hsize=0.4\hsize}X
   >{\centering\arraybackslash\hsize=0.6\hsize}X
   >{\centering\arraybackslash\hsize=0.6\hsize}X
   >{\centering\arraybackslash\hsize=0.6\hsize}X}
\toprule
\textbf{Test Name} & \textbf{Yes} & \textbf{Partial} & \textbf{No} & \textbf{Start Failed} & \textbf{Accuracy} & \textbf{Appearance Score} \\
\midrule \midrule
\multicolumn{7}{c}{\textbf{OpenHands}}\\ \midrule
Claude-3.5-Sonnet    & 18.1 &  8.3 & 58.6 & 15.0 & 22.3 & 2.6 \\
DeepSeek-R1          &  8.5 &  3.4 & 60.4 & 27.7 & 10.2 & 1.4 \\
DeepSeek-V3          &  7.4 &  3.2 & 73.9 & 15.5 &  9.0 & 1.5 \\
\midrule \midrule
\multicolumn{7}{c}{\textbf{Aider}}\\ \midrule
Claude-3.5-Sonnet    & 19.9 &  5.9 & 42.0 & 32.1 & 22.9 & 1.9 \\
DeepSeek-R1          & 23.3 &  8.7 & 44.5 & 23.5 & 27.7 & 2.7 \\
DeepSeek-V3          & 12.5 &  3.1 & 54.3 & 30.1 & 14.1 & 1.3 \\
\midrule \midrule
\multicolumn{7}{c}{\textbf{Bolt.diy}}\\ \midrule
Claude-3.5-Sonnet    & 22.6 &  7.6 & 64.1 &  5.7 & 26.4 & 3.0 \\
DeepSeek-R1          & 24.7 &  6.2 & 64.3 &  4.8 & 27.8 & 2.5 \\
DeepSeek-V3          & 18.5 &  4.5 & 73.9 &  3.1 & 20.8 & 2.0 \\
GPT-4o               & 10.4 &  4.8 & 64.5 & 20.4 & 12.8 & 1.5 \\
o3-mini              & 17.9 &  3.4 & 40.0 & 38.6 & 19.6 & 1.6 \\
Qwen2.5-Coder-32B-Inst.    &  8.2 &  2.6 & 81.8 &  7.4 &  9.5 & 1.1 \\
Qwen2.5-72B-Inst. & 12.1 &  3.6 & 80.7 &  3.7 & 13.8 & 1.4 \\
WebGen-LM-7B         & 24.9 &  7.1 & 68.0 &  0.0 & 28.4 & 2.5 \\
WebGen-LM-14B        & 25.0 &  8.7 & 66.3 &  0.0 & 29.4 & 2.5 \\
WebGen-LM-32B        & 34.2 &  8.0 & 57.8 &  0.0 & 38.2 & 2.8 \\
\midrule \midrule
\multicolumn{7}{c}{\textbf{WebGen-Agent}}\\ \midrule
\multicolumn{7}{c}{\textbf{Proprietary Models}}\\ \midrule
Claude-3.5-Sonnet            & 45.6 & 12.7 & 40.6 & 1.1 & 51.9 & 3.9 \\
DeepSeek-R1                  & 40.2 & 12.4 & 45.9 & 1.5 & 46.4 & 3.8 \\
DeepSeek-V3                  & 46.1 & 13.1 & 40.6 & 0.2 & 52.6 & 3.8 \\
o3                            & 45.7 & 11.9 & 41.6 & 0.8 & 51.7 & 3.5 \\
Gemini-2.5-Pro               & 44.5 & 12.7 & 39.4 & 3.4 & 50.9 & 3.8 \\
Claude-4-Sonnet              & 48.8 & 15.3 & 33.4 & 2.5 & 56.5 & 4.1 \\
Qwen3-Coder-480B-A35B-Inst.  & 50.5	&15.3	&34.2	&0.0	&\textbf{58.2}	&\textbf{4.3} \\
\midrule
\multicolumn{7}{c}{\textbf{Open-Source Models (30B–72B)}}\\ \midrule
Qwen2.5-Coder-32B-Inst.      & 26.7 & 10.5 & 60.3 & 2.5 & 32.0 & 3.3 \\
Qwen3-Coder-30B-A3B-Inst.    & 45.7 & 14.1 & 40.2 & 0.0 & \textbf{52.8} & \textbf{4.0} \\
Qwen2.5-72B-Inst.         & 29.1	&13.8	&57.2	&0.0	&35.9	& 3.4 \\
\midrule
\multicolumn{7}{c}{\textbf{Open-Source Models (7B–8B)}}\\ \midrule
Qwen2.5-Coder-7B-Inst.                       & 10.0 &  4.8 & 60.9 & 24.3 & 12.4 & 1.6 \\
WebGenAgent-LM-7B-SFT             & 33.8 & 10.2 & 56.0 &  0.0 & 38.9 & 3.4 \\
WebGenAgent-LM-7B-Step-GRPO     & 40.2 & 10.5 & 49.3 &  0.0 & \textbf{45.4} & \textbf{3.7} \\
\midrule
Qwen3-8B                                     & 29.5 &  9.1 & 61.4 &  0.0 & 34.1 & 3.2 \\
WebGenAgent-LM-8B-SFT                             & 32.8 & 11.6 & 55.6 &  0.0 & 38.6 & 3.4 \\
WebGenAgent-LM-8B-Step-GRPO    &37.4	&12.1	&50.5	&0.0	&\textbf{43.4}	&\textbf{3.6}  \\
\bottomrule
\end{tabularx}
\label{tab:main_results}
\end{table}

\paragraph{Benchmark Dataset and Baselines.} We evaluate WebGen-Agent using WebGen-Bench~\citep{lu2025webgenbenchevaluatingllmsgenerating}, a benchmark containing 101 website-generation instructions in natural language and 647 GUI-agent test cases, covering a wide range of web applications. Following~\cite{lu2025webgenbenchevaluatingllmsgenerating}, we use Qwen2.5-VL-32B-Instruct~\citep{bai2025qwen25vltechnicalreport} in functional testing and GPT-4o~\citep{hurst2024gpt} in appearance evaluation. We compare WebGen-Agent with three other popular code agents: OpenHands~\citep{wang2024openhands}, Aider~\citep{aiderai2024aider}, and Bolt.diy~\citep{stackblitzlabs2024bolt}. We present the results of OpenHands and Aider in combination with DeepSeek-V3~\citep{liu2024deepseek}, Claude-3.5-Sonnet~\citep{anthropic2024claude}, and DeepSeek-R1~\citep{guo2025deepseek}, as well as the results of Bolt.diy with DeepSeek-V3~\citep{liu2024deepseek}, Claude-3.5-Sonnet~\citep{anthropic2024claude}, DeepSeek-R1~\citep{guo2025deepseek}, GPT-4o~\citep{hurst2024gpt}, o3-mini~\citep{openai2025o3mini}, Qwen2.5-Coder-32B~\citep{hui2024qwen2}, Qwen2.5-72B-Instruct~\citep{yang2024qwen2}, WebGen-LM-7B, WebGen-LM-14B, and WebGen-LM-32B~\citep{lu2025webgenbenchevaluatingllmsgenerating}. The values are taken from~\citep{lu2025webgenbenchevaluatingllmsgenerating}.

\paragraph{Models and WebGen-Agent Inference Settings.} We evaluate WebGen-Agent using a wide range of proprietary and open-source models as coding LLMs. The proprietary models we tested include Claude-3.5-Sonnet~\citep{anthropic2024claude}, DeepSeek-R1~\citep{guo2025deepseek}, DeepSeek-V3~\citep{liu2024deepseek}, o3~\citep{openai2025o3}, Claude-4-Sonnet~\citep{anthropic2024claude4}, Gemini-2.5-Pro~\citep{comanici2025gemini}, and Qwen3-Coder-480B-A35B-Instruct~\citep{yang2025qwen3}. The smaller open-source models we tested include Qwen2.5-Coder-32B-Instruct~\citep{hui2024qwen2}, Qwen3-Coder-30B-A3B-Instruct~\citep{yang2025qwen3}, Qwen2.5-72B-Instruct~\citep{yang2024qwen2}, Qwen2.5-Coder-7B-Instruct~\citep{hui2024qwen2}, and Qwen3-8B~\citep{yang2025qwen3}, as well as 7B and 8B WebGenAgent-LM models trained with SFT and Step-GRPO. The maximum number of iterations is set to 20, and the model temperature is set to 0.5. We use Qwen2.5-VL-32B-Instruct as the feedback VLM for screenshot and GUI-agent testing in all the experiments. Analysis of the maximum iteration number is presented in Appendix~\ref{sec:maximum_iteration_number}.

\paragraph{Training Settings.} We first fine-tune Qwen2.5-Coder-7B-Instruct and Qwen3-8B on approximately seven hundred WebGen-Agent trajectories collected from DeepSeek-V3 for one epoch with a learning rate of 4e-5 and a batch size of 32. This results in the models WebGenAgent-LM-7B-SFT and WebGenAgent-LM-8B-SFT, which serve as a warm start for Step-GRPO. We then train these SFT models using Step-GRPO on five hundred website generation instructions randomly sampled from WebGen-Instruct for one epoch, resulting in the final models WebGenAgent-LM-7B-Step-GRPO and WebGenAgent-LM-8B-Step-GRPO. The learning rate is set to 1e-6 with a batch size of 16. For each instruction, we sample 5 outputs. Ambiguous or underspecified instructions are manually filtered out. We observe that this relatively small number of high-quality instructions is sufficient for Step-GRPO training, likely due to the reliable step-level feedback from screenshots and the GUI agent. Training on more samples is costly and does not yield noticeable gains.

\paragraph{Results.}
The WebGen-Agent test results are presented in Tab.~\ref{tab:main_results}. Based on the results, we make the following observations: 
(1) WebGen-Agent demonstrates superior performance across various proprietary models compared to other code agent systems. On Claude-3.5-Sonnet, DeepSeek-R1, and DeepSeek-V3, WebGen-Agent significantly outperforms OpenHands, Aider, and Bolt.diy when using the same model. Across all seven proprietary models from five different providers, WebGen-Agent achieves consistently high performance, demonstrating the generalizability of the method. Qwen3-Coder-480B-A35B-Instruct achieves the highest accuracy of 58.2\% and an appearance score of 4.3.
(2) With 30B–72B sized open-source models, WebGen-Agent also achieves high performance. On Qwen2.5-Coder-32B-Instruct and Qwen2.5-72B-Instruct, WebGen-Agent outperforms the previous state-of-the-art, Bolt.diy, by 22.5\% and 22.1\% in accuracy, and by 2.2 and 2.0 in appearance scores, respectively. Qwen3-Coder-30B-A3B-Instruct achieves the best performance among 30B–72B models, with 52.8\% accuracy and an appearance score of 4.0.
(3) Step-GRPO with Screenshot and GUI-agent Feedback significantly improves the performance of Qwen2.5-Coder-7B-Instruct and Qwen3-8B. For Qwen2.5-Coder-7B-Instruct, SFT improves accuracy from 12.4\% to 38.9\% and the appearance score from 1.6 to 3.4; Step-GRPO further improves accuracy from 38.9\% to 45.4\% and the appearance score from 3.4 to 3.7. For Qwen3-8B, SFT improves accuracy from 34.1\% to 38.6\% and the appearance score from 3.2 to 3.4; Step-GRPO further improves accuracy from 38.6\% to 43.4\% and the appearance score from 3.4 to 3.6. Qualitative analysis of SFT and Step-GRPO's effect in improving the performance is presented in Appendix~\ref{sec:sft_step_grpo_qualitative_analysis}. These results demonstrate the effectiveness of our training method in improving both the functionality and appearance of the generated websites. Categorical results are presented in Tab.~\ref{tab:categorical_results} of Appendix~\ref{sec:categorical_results}.

\subsection{Ablation Studies}

\begin{table}[t]\fontsize{9}{10}\selectfont
\centering
\caption{Ablation study on the WebGen‐Agent workflow.  
The configuration starts from execution-only and incrementally adds capabilities.}
\begin{tabularx}{\textwidth}
  {>{\raggedright\arraybackslash\hsize=3.0\hsize}X
   >{\centering\arraybackslash\hsize=0.3\hsize}X
   >{\centering\arraybackslash\hsize=0.3\hsize}X
   >{\centering\arraybackslash\hsize=0.3\hsize}X
   >{\centering\arraybackslash\hsize=0.3\hsize}X
   >{\centering\arraybackslash\hsize=0.5\hsize}X
   >{\centering\arraybackslash\hsize=0.5\hsize}X}
\toprule
\textbf{Test Name} & \textbf{Yes} & \textbf{Partial} & \textbf{No} &
\textbf{Start Failed} & \textbf{Accuracy} & \textbf{Appearance Score} \\
\midrule
Execution‐only                                                        & 39.7 & 12.4 & 43.3 & 4.6 & 45.9 & 3.0 \\
Screenshot                                              & 41.3 & 10.7 & 45.9 & 2.2 & 46.6 & 3.6 \\
Screenshot+GUI-agent                                & 43.0 & 13.9 & 41.3 & 1.9 & 49.9 & 3.4 \\
Screenshot+GUI-agent+Backtrack                  & 45.6 & 11.1 & 43.1 & 0.2 & 51.2 & 3.7 \\
\textbf{Screenshot+GUI-agent+Backtrack+Select-best}                & 46.1 & 13.1 & 40.6 & 0.2 & \textbf{52.6} & \textbf{3.8} \\
\bottomrule
\end{tabularx}
\label{tab:inference_ablation}
\end{table}

\paragraph{Analysis of the WebGen-Agent Workflow.}
We analyze various design choices in the WebGen-Agent workflow in Tab.~\ref{tab:inference_ablation}. We incrementally add the designs, starting from using only the code execution response messages $\mathcal{O}$ (“Execution-only”), then gradually adding screenshot feedback $\mathcal{F}_\text{shot}$ (“Screenshot”), GUI-agent testing feedback $\mathcal{F}_\text{gui}$ (“Screenshot+GUI-agent”), the backtracking mechanism (“Screenshot+GUI-agent+Backtrack”), and finally the select-best mechanism (“Screenshot+GUI-agent+Backtrack+Select-best”), which makes up the full WebGen-Agent workflow. As shown in Tab.~\ref{tab:inference_ablation}, each of these designs yields notable gains in accuracy and appearance. The GUI-agent testing contributes the largest accuracy gain of 3.3\%, showing its effectiveness in guiding the functionality of the generated websites. The addition of screenshot feedback greatly improves the appearance score, raising it from 3.0 to 3.6, demonstrating its effect in enhancing website appearance. Adding GUI-agent testing slightly impairs the appearance score, likely because modifying the codebase for functional fulfillment sometimes damages the website appearance or causes errors. This negative effect is mitigated by the addition of the backtracking and select-best mechanisms. Qualitative analysis of the effect of screenshot and GUI-agent feedback is provided in Appendix~\ref{sec:workflow_qualitative_analysis}. Also, as shown in Tab.~\ref{tab:gui_instruction_scores} of Appendix~\ref{sec:gui_instruction_comprehensiveness}, a manual inspection indicates that 98.3\% of the sampled instructions achieve high coverage of the requirements.

\begin{table}[t]\fontsize{9}{10}\selectfont
\centering
\caption{Training strategy ablation on the Qwen2.5-Coder-7B-Instruct model. The configuration starts from the raw model and successively introduces supervised fine-tuning (SFT) and various reinforcement-learning variants.}
\begin{tabularx}{\textwidth}
  {>{\raggedright\arraybackslash\hsize=3.0\hsize}X
   >{\centering\arraybackslash\hsize=0.3\hsize}X
   >{\centering\arraybackslash\hsize=0.4\hsize}X
   >{\centering\arraybackslash\hsize=0.3\hsize}X
   >{\centering\arraybackslash\hsize=0.3\hsize}X
   >{\centering\arraybackslash\hsize=0.5\hsize}X
   >{\centering\arraybackslash\hsize=0.6\hsize}X}
\toprule
\textbf{Test Name} & \textbf{Yes} & \textbf{Partial} & \textbf{No} &
\textbf{Start Failed} & \textbf{Accuracy} & \textbf{Appearance Score} \\
\midrule
No Additional Training                                   & 10.0 &  4.8 & 60.9 & 24.3 & 12.4 & 1.6 \\
\midrule
SFT for 1 Epoch                                          & 33.8 & 10.2 & 56.0 &  0.0 & 38.9 & 3.4 \\
SFT for 2 Epochs                                         &    32.1	&14.2	&53.5	&0.2	&39.3  &3.4  \\
\midrule
Naive Outcome GRPO                                       & 38.0 &  9.0 & 53.0 &  0.0 & 42.5 & 3.5 \\
Step-GRPO w/ Cumulative Advantage                        & 32.6 & 12.2 & 55.2 &  0.0 & 38.7 & 3.5 \\
\midrule
Step-GRPO w/ Screenshot Reward Only                      & 34.9 & 10.5 & 53.9 &  0.6 & 40.2 & 3.5 \\
Step-GRPO w/ GUI-agent Reward Only                       & 34.8 & 11.3 & 53.6 &  0.3 & 40.4 & 3.4 \\
\midrule
\textbf{Step-GRPO w/ Screenshot+GUI-agent (ours)}          & 40.2 & 10.5 & 49.3 &  0.0 & \textbf{45.4} & \textbf{3.7} \\
\bottomrule
\end{tabularx}
\label{tab:training_ablation}
\end{table}

\begin{figure*}[t]
    \centering
    \includegraphics[width=0.6\textwidth]{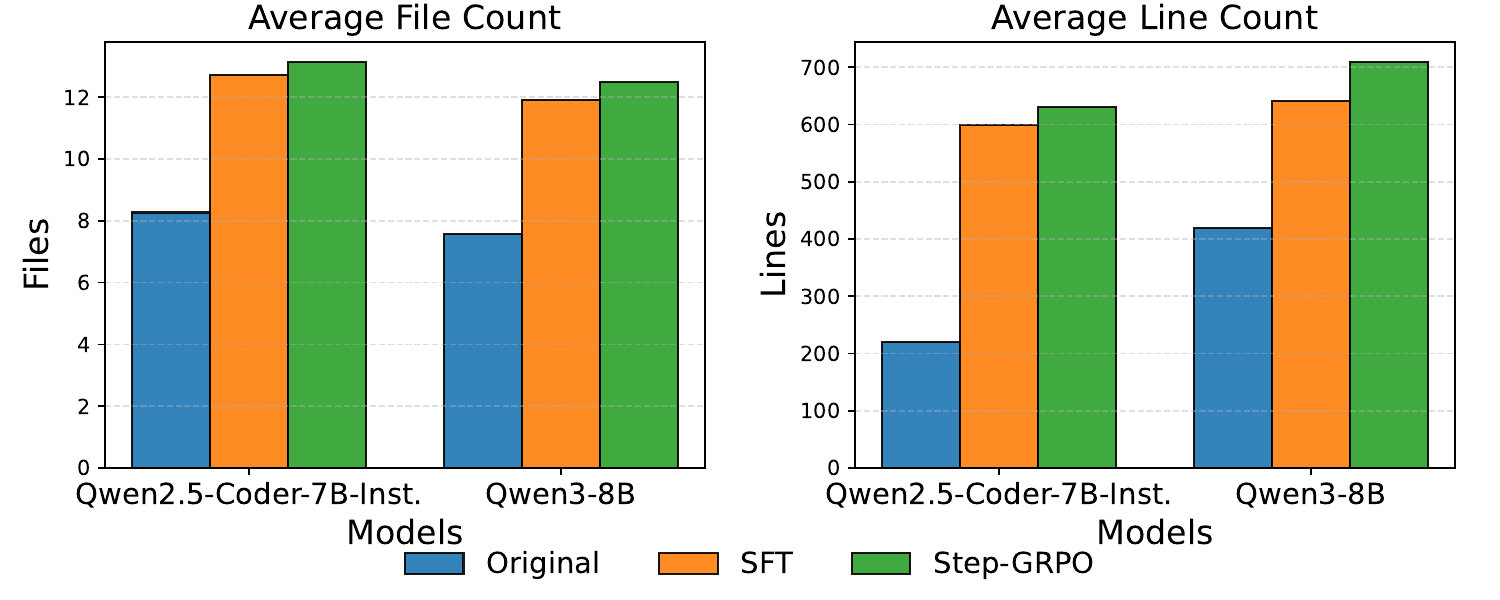}
    \caption{Comparison of the average file count and average line count among the original, SFT, and Step-GRPO models for Qwen2.5-Coder-7B-Instruct and Qwen3-8B.}
    
\label{fig:files_lines_number}
\end{figure*}

\paragraph{Analysis of Step-GRPO with Screenshot and GUI-agent Feedback.}
We analyze the design choices in the Step-GRPO with Screenshot and GUI-agent Feedback training process in Tab.~\ref{tab:training_ablation}. The first line shows the result of the Qwen2.5-Coder-7B-Instruct model with no additional training. The analysis based on Tab.~\ref{tab:training_ablation} is as follows:
(1) The second and third lines present SFT training for one epoch and two epochs, showing that training for two epochs does not notably improve performance compared to training for only one epoch. Therefore, we train for only one epoch in the SFT stage.
(2) The fourth and fifth lines show the results of using naive outcome GRPO and Step-GRPO with cumulative advantage. The rewards in these two variants are the same as in our final design ($\textit{Score}_{\text{shot}} + \textit{Score}_{\text{gui}}$); only the advantage computation method differs. Naive outcome GRPO uses the maximum value of the step-level rewards in a trajectory as the outcome reward, setting the advantages to the normalized outcome rewards. Step-GRPO with cumulative advantage calculates the advantage of each token as the sum of the normalized rewards from the subsequent steps, as introduced in ~\cite{shao2024deepseekmathpushinglimitsmathematical}. Both GRPO advantage computation variants perform notably worse than our final Step-GRPO setting.
(3) The sixth and seventh lines present the results of using only the screenshot scores ($\textit{Score}_{\text{shot}}$) or only the GUI-agent testing scores ($\textit{Score}_{\text{gui}}$) as the rewards. Both are lower than using $\textit{Score}_{\text{shot}} + \textit{Score}_{\text{gui}}$, demonstrating the necessity of incorporating both types of feedback.
We also gather statistics on the average file count and average line count for the Original, SFT, and Step-GRPO models, as shown in Fig.~\ref{fig:files_lines_number}. For both Qwen2.5-Coder-7B-Instruct and Qwen3-8B, the average file count and average line count consistently increase after SFT and Step-GRPO training. This shows that both the SFT and Step-GRPO stages increase the complexity of the generated websites, which is consistent with their improved performance.

\begin{table}[t]\fontsize{9}{10}\selectfont
\centering
\caption{Impact of the feedback VLM on the performance of WebGen-Agent on WebGen-Bench. The highest Accuracy and Appearance Score are highlighted in \textbf{bold}.}
\begin{tabularx}{\textwidth}
  {>{\raggedright\arraybackslash\hsize=1.5\hsize}X
   >{\raggedright\arraybackslash\hsize=1.6\hsize}X
   >{\centering\arraybackslash\hsize=0.3\hsize}X
   >{\centering\arraybackslash\hsize=0.3\hsize}X
   >{\centering\arraybackslash\hsize=0.3\hsize}X
   >{\centering\arraybackslash\hsize=0.3\hsize}X
   >{\centering\arraybackslash\hsize=0.5\hsize}X
   >{\centering\arraybackslash\hsize=0.5\hsize}X}
\toprule
\textbf{Coding LLM} & \textbf{Feedback VLM} &
\textbf{Yes} & \textbf{Partial} & \textbf{No} &
\textbf{Start Failed} & \textbf{Accuracy} & \textbf{Appearance Score} \\
\midrule\midrule
Qwen2.5-VL-32B-Inst. & Qwen2.5-VL-32B-Inst. &  4.5 &  2.2 & 78.8 & 14.5 &  5.6 & 1.3 \\
DeepSeek-V3          & GPT-4o               & 46.4 & 11.4 & 42.0 &  0.2 & 52.1 & 3.6 \\
\textbf{DeepSeek-V3}          & \textbf{Qwen2.5-VL-32B-Inst.} & 46.1 & 13.1 & 40.6 &  0.2 & \textbf{52.6} & \textbf{3.8} \\
\bottomrule
\end{tabularx}
\label{tab:vlm_ablation}
\end{table}

\paragraph{Analysis of the Coding LLM and Feedback VLM.} We analyze the choice of the coding LLM and feedback VLM in Tab.~\ref{tab:vlm_ablation}. In our experiments, we use a relatively small and inexpensive VLM, Qwen2.5-VL-32B-Instruct, to provide screenshot and GUI-agent testing feedback, while employing a strong LLM capable of generating high-quality code, such as DeepSeek-V3, as the coding LLM. As shown in the second row of Tab.~\ref{tab:vlm_ablation}, replacing Qwen2.5-VL-32B-Instruct with a proprietary VLM, GPT-4o, as the feedback VLM does not notably improve the accuracy or the appearance score. This demonstrates that Qwen2.5-VL-32B-Instruct is already sufficient for providing accurate screenshot and GUI-agent testing feedback, while being more cost-effective than proprietary VLMs. As shown in the first row of Tab.~\ref{tab:vlm_ablation}, replacing DeepSeek-V3 with Qwen2.5-VL-32B-Instruct results in significantly worse performance, indicating that the coding LLM cannot be replaced by smaller open-source VLMs. The design choice of decoupling the coding LLM and feedback VLM ensures that code is generated by a strong LLM to maintain quality, while screenshot and GUI-agent testing feedback is handled by a smaller open-source VLM for cost efficiency. Further analysis of the accuracy of the screenshot and GUI-agent scores provided by the feedback VLM is included in Tab.~\ref{tab:score_accuracies} of Appendix~\ref{sec:score_acc}, demonstrating the reliability of the scores.

\section{Related Work}

\paragraph{Visual Code Generation.} Code generation that is associated with visual effects exists in a wide range of application scenarios, such as web page development~\citep{lu2025webgenbenchevaluatingllmsgenerating,xu2025webbenchllmcodebenchmark} and GitHub-issue fixing~\citep{yang2024swebenchmultimodalaisystems, guo2025omnigirlmultilingualmultimodalbenchmark}. Previous work has proposed various ways to treat visual elements in code generation and other reasoning-intensive tasks~\citep{su2025thinkingimagesmultimodalreasoning}, such as generating code to represent images in problem statements~\citep{huang2025seeingfixingcrossmodalreasoning, wang2025mathcodervlbridgingvisioncode} and using natural language to describe images\citep{zhang2024codevissueresolvingvisual}. We also apply natural language descriptions when providing screenshot feedback. More related to our work, a line of studies\citep{guo2024iwbenchevaluatinglargemultimodal,si2025design2codebenchmarkingmultimodalcode,yun2024web2codelargescalewebpagetocodedataset,beltramelli2017pix2codegeneratingcodegraphical,sun2025fullfrontbenchmarkingmllmsfrontend,Gui_2025,laurençon2024unlockingconversionwebscreenshots,wan2024mrwebexplorationgeneratingmultipage} explores MLLMs’ ability to reconstruct single-file HTML code from webpage screenshots. Other studies benchmark MLLMs’ performance in implementing interactive elements in existing web projects~\citep{xiao2025interaction2codebenchmarkingmllmbasedinteractive} or performing web development tasks in a pre-defined sequential manner with detailed technical settings~\citep{xiao2025designbenchcomprehensivebenchmarkmllmbased,xu2025webbenchllmcodebenchmark}. The web development tasks in these works are often solved in a single HTML file~\citep{zhang2025artifactsbenchbridgingvisualinteractivegap} or contain rigid pipelines~\citep{xu2025webbenchllmcodebenchmark}, which are more suitable for testing MLLMs rather than code agents for end-to-end, repository-level website development, as proposed in our work. Therefore, we evaluate our agent workflow with WebGen-Bench~\citep{lu2025webgenbenchevaluatingllmsgenerating}, which measures a code agent’s ability to create multi-file website codebases from scratch and includes diverse website generation instructions.

\paragraph{Code Agents.} Equipped with various tools and powered by LLMs~\citep{soni2025codingagentsmultimodalbrowsing, yao2023reactsynergizingreasoningacting, zhang2024codeagentenhancingcodegeneration}, code agents can perform a variety of tasks, such as developing websites~\citep{lu2025webgenbenchevaluatingllmsgenerating} and fixing GitHub issues~\citep{jimenez2024swebenchlanguagemodelsresolve, yang2024sweagentagentcomputerinterfacesenable}. Some code agents specialize in a specific field, such as bug fixing~\citep{zhang2024autocoderoverautonomousprogramimprovement} or machine learning~\citep{jiang2025aideaidrivenexplorationspace}. Similar to our work, Bolt.diy~\citep{stackblitzlabs2024bolt} specializes in multi-file website generation. Others, such as OpenHands~\citep{wang2024openhands} and Aider~\citep{aiderai2024aider}, are general-purpose code agents that are not limited to a single field, though their performance on a specific task might not match that of specialist code agents~\citep{lu2025webgenbenchevaluatingllmsgenerating}. Our WebGen-Agent is a code agent specializing in end-to-end website generation, with screenshot feedback and GUI-agent testing features specifically designed for this task, achieving state-of-the-art performance.

\paragraph{Fine-tuning and Reinforcement Learning for Agents.}
Supervised fine-tuning~\citep{pan2025trainingsoftwareengineeringagents,yang2025swesmithscalingdatasoftware} and reinforcement learning~\citep{dong2025agenticreinforcedpolicyoptimization,qian2025toolrlrewardtoollearning} are two methods widely used to improve the agentic and tool-calling abilities of LLMs. In the field of code agents, various works~\citep{pan2025trainingsoftwareengineeringagents,yang2025swesmithscalingdatasoftware,zhang2025sealignalignmenttrainingsoftware,wang2025swedevbuildingsoftwareengineering,ma2024lingmaswegptopendevelopmentprocesscentric,xie2025swefixertrainingopensourcellms,jain2025r2egymproceduralenvironmentshybrid,guo2025swefactoryautomatedfactoryissue,ma2025thinkinglongerlargerenhancing} leverage supervised fine-tuning combined with software engineering data synthesis and rejection sampling to improve the performance of open-source models. Similar to these works, we also use rejection sampling and supervised fine-tuning in the warm-up stage before the GRPO training. Other works use reinforcement learning with rewards acquired through comparison with the ground truth~\citep{wei2025swerladvancingllmreasoning,ma2025toolintegratedreinforcementlearningrepo,zhuang2025workforceagentr1incentivizingreasoningcapability}, determined by the code execution output~\citep{gehring2025rlefgroundingcodellms,ma2025sorftissueresolvingsubtaskoriented,golubev2025traininglongcontextmultiturnsoftware}, or dependent on task success~\citep{wei2025webagentr1trainingwebagents,lu2025arpoendtoendpolicyoptimizationgui,chen2025r1codeinterpretertrainingllmsreason}. These works either use outcome supervision, which provides sparse training signals, or require detailed ground truth to provide step supervision, which is rigid and difficult to obtain. In contrast to these methods, our work leverages screenshot and GUI-agent testing scores at each step—which are inherent in the WebGen-Agent pipeline—to provide accurate step-level supervision in GRPO training.

\section{Conclusion}

In this paper, we introduce WebGen-Agent, a code agent that leverages screenshot and GUI-agent testing feedback, combined with backtracking and select-best mechanisms, to iteratively generate websites with appealing appearance and smooth functionality. We also propose Step-GRPO with Screenshot and GUI-agent Feedback, which leverages inherent screenshot and GUI-agent testing scores to provide step-level supervision in the GRPO training process. Testing WebGen-Agent on WebGen-Bench shows significant improvements across a wide range of proprietary and open-source LLMs compared to other code agent systems. WebGen-Agent with Qwen3-Coder-480B-A35B-Instruct achieves the best performance, with an accuracy of 58.2\% and an appearance score of 4.3. Training Qwen2.5-Coder-7B-Instruct and Qwen3-8B first with supervised fine-tuning and then with Step-GRPO with Screenshot and GUI-agent Feedback notably improves accuracies and appearance scores, demonstrating the effectiveness of our training approach.

\section{Reproducibility Statement}

To ensure reproducibility, we release the WebGen-Agent workflow code, along with the training code and data for Step-GRPO with Screenshot and GUI-Agent Feedback, as well as the weights of the WebGenAgent-LM models. The complete code base and datasets are provided in the supplementary material accompanying this paper. Details of the agent workflow and of all prompts used to deliver multi-level feedback are presented in Section~\ref{sec:webgen-agent-workflow}, Appendix~\ref{sec:webgen_agent_algorithm}, and Appendix~\ref{sec:webgen_agent_prompts}. The training procedure for Step-GRPO with Screenshot and GUI-Agent Feedback is described in Section~\ref{sec:step-grpo} and Section~\ref{sec:main-results}. We also report a manual inspection of the screenshot and GUI-agent scores in Appendix~\ref{sec:score_acc}, and we assess the comprehensiveness of the GUI-Agent testing instructions in Appendix~\ref{sec:gui_instruction_comprehensiveness}. Collectively, these resources ensure that our findings are transparent, robust, and independently verifiable.

\bibliography{iclr2026_conference}
\bibliographystyle{iclr2026_conference}

\appendix

\section{Limitations and Future Work}

WebGen-Agent is specifically designed to generate websites based on natural language instructions from non-expert users. We do not consider website response speed or complex network conditions when generating and evaluating the websites; these are interesting questions for future work. In the supervised fine-tuning and Step-GRPO experiments, we train only 7B- and 8B-parameter models due to limited computing power and GPU memory, as Step-GRPO training would take more than 24 hours on 16 NVIDIA A800 GPUs, and we currently do not have enough GPUs to train larger models. The results on the 7B and 8B models show great potential for our method, and we plan to apply our training approach to 30B–72B models in the future.

\section{WebGen-Agent Algorithm}
\label{sec:webgen_agent_algorithm}

Algorithm~\ref{alg:webgen_agent} demonstrates the WebGen-Agent inference workflow in detail. Algorithms~\ref{alg:select_best} and~\ref{alg:truncate} are two helper functions for Algorithm~\ref{alg:webgen_agent}, presented separately for clarity.
\begin{algorithm}[t]
  \caption{WebGen-Agent}
  \label{alg:webgen_agent}
  \small
  \begin{algorithmic}[1]
    \Require Initial instruction $\mathcal{I}$, maximum steps $T$
    \Ensure  Final codebase $\mathcal{C}^\star$                
    \State $\mathcal{T}\ \gets [\,\mathcal{I}\,]$               \Comment{trajectory: instruction, \textbf{edit}, feedback, …}
    \State $\textit{Steps}\ \gets \emptyset$                    \Comment{archive of step snapshots}
    \State $\mathcal{C}\ \gets \emptyset$                       \Comment{current \emph{codebase}}
    \State $t \gets 1$, \; $\textit{consecErr} \gets 0$
    \While{$t \le T$}
        \State $\Delta\mathcal{C}_t \gets \textsc{GenerateEdit}(\mathcal{T})$
        \State $\mathcal{T} \mathrel{+\!\!=} \Delta\mathcal{C}_t$
        \State $\mathcal{C} \gets \textsc{ApplyEdit}(\mathcal{C},\Delta\mathcal{C}_t)$
        \State $\mathcal{O} \gets \textsc{Execute}(\mathcal{C})$
        \If{$\mathcal{O} = \textit{error}$}
            \State $\mathcal{T} \mathrel{+\!\!=} \mathcal{O}$
            \State $\textit{consecErr} \gets \textit{consecErr}+1$
            \If{$\textit{consecErr}=5$}
                \State $\bigl\langle t^\star,\mathcal{C}^\star,*,*\bigr\rangle
                       \gets \textsc{SelectBestStep}(\textit{Steps})$
                \State $\mathcal{C} \gets \mathcal{C}^\star$               \Comment{restore codebase}
                \State $\mathcal{T} \gets \textsc{Truncate}(\mathcal{T},t^\star)$
                \State $t \gets t^\star + 1$, \; $\textit{consecErr} \gets 0$
            \Else
                \State $t \gets t+1$
            \EndIf
            \State \textbf{continue}
        \Else
            \State $\textit{consecErr} \gets 0$
        \EndIf
        \State $\textit{img} \gets \textsc{Screenshot}(\mathcal{C})$
        \State $\bigl\langle\textit{desc},\textit{sugg}_{\text{shot}},\textit{score}_{\text{shot}}\bigr\rangle \gets
               \textsc{VLM\_Judge}(\textit{img})$
        \State $\mathcal{T} \mathrel{+\!\!=}
               \langle\textit{desc},\textit{sugg}_{\text{shot}}\rangle$
        \State $\textit{goNext} \gets \textsc{AgentDecision}(\mathcal{T})$
        \If{\textbf{not} $\textit{goNext}$}
            \State $t \gets t+1$;\; \textbf{continue}
        \EndIf
        \State $\bigl\langle\textit{pass},\textit{sugg}_{\text{gui}},\textit{score}_{\text{gui}}) \gets
               \textsc{GUI\_Agent}(\mathcal{C}\bigr\rangle$
        \State $\mathcal{T} \mathrel{+\!\!=}
               \langle\textit{pass},\textit{sugg}_{\text{gui}}\rangle$
        \State $\textit{Steps} \mathrel{+\!\!=}
               \bigl\langle t,\mathcal{C},\textit{score}_{\text{shot}},\textit{score}_{\text{gui}}\bigr\rangle$
        \If{\textit{pass}}
            \State \textbf{break}
        \Else
            \State $t \gets t+1$
        \EndIf
    \EndWhile
    \State $\bigl\langle *,\mathcal{C}^\star,*,*\bigr\rangle \gets \textsc{SelectBestStep}(\textit{Steps})$
    \State \Return $\mathcal{C}^\star$
  \end{algorithmic}
\end{algorithm}

\begin{algorithm}[t]
  \caption{\textsc{SelectBestStep}}
  \label{alg:select_best}
  \small
  \begin{algorithmic}[1]
    \Require $\textit{Steps}=\{\langle t,\mathcal{C},\textit{score}_{\text{shot}},\textit{score}_{\text{gui}}\rangle\}$
    \State $g_{\max} \gets \max_{s\in\textit{Steps}} \textit{score}_{\text{gui}}$
    \State $\mathcal{S}_g \gets \{s\mid \textit{score}_{\text{gui}} = g_{\max}\}$
    \State \Return $\displaystyle\arg\max_{s\in\mathcal{S}_g} \textit{score}_{\text{shot}}$
  \end{algorithmic}
\end{algorithm}

\begin{algorithm}[t]
  \caption{\textsc{Truncate}}
  \label{alg:truncate}
  \small
  \begin{algorithmic}[1]
    \Require Trajectory $\mathcal{T}$, step id $t^\star$
    \State \Return prefix of $\mathcal{T}$ ending just after the edit and feedback of step $t^\star$
  \end{algorithmic}
\end{algorithm}

\section{WebGen-Agent Prompts}
\label{sec:webgen_agent_prompts}

The prompts for acquiring screenshot and GUI-agent testing feedback are presented in Fig.~\ref{fig:prompt_screenshot_desc}, Fig.~\ref{fig:webpage_design_prompt}, Fig.~\ref{fig:gui_agent_trigger_prompt}, and Fig.~\ref{fig:gui_agent_testing_prompt}.
 
\begin{figure}
  \begin{tcolorbox}[colback=blue!5!white,colframe=blue!75!black]
    \begin{small}
      \textbf{Prompt:}\\
      You are given a single website screenshot as input.

      \textbf{Task}
      \begin{enumerate}[leftmargin=1.5em,itemsep=0.3em,topsep=0.3em]
        \item Examine the screenshot closely for any rendering or runtime errors (e.g., ``404 Not Found'', stack traces, missing styles, blank areas).
        \item Decide whether the screenshot \textit{shows a rendering or runtime error}.
              \begin{itemize}[leftmargin=1.5em,itemsep=0.3em,topsep=0.3em]
                \item If \textbf{yes}, set ``\texttt{is\_error}'': \texttt{true}, extract or paraphrase the visible error message(s) into ``\texttt{error\_message}'', and leave ``\texttt{screenshot\_description}'' empty.
                \item If \textbf{no}, set ``\texttt{is\_error}'': \texttt{false}, leave ``\texttt{error\_message}'' as an empty string (``''), and write a concise but thorough ``\texttt{screenshot\_description}'' that covers:
                      \begin{itemize}[leftmargin=1.5em,itemsep=0.2em,topsep=0.2em]
                        \item Overall layout (e.g., header/sidebar/footer, grid, flex, single-column).
                        \item Key UI components (navigation bar, buttons, forms, images, cards, tables, modals, etc.).
                        \item Color scheme and visual style (dominant colors, light/dark theme, gradients, shadows).
                        \item Visible content and text (headings, labels, sample data).
                        \item Notable design details (icons, spacing, font style) that help someone understand what the page looks like).
                      \end{itemize}
              \end{itemize}
        \item Suggest ways to improve the appearance of the website, for example:
              \begin{itemize}[leftmargin=1.5em,itemsep=0.3em,topsep=0.3em]
                \item Separate incorrectly overlapping components.
                \item Adjust layout to avoid large blank areas.
                \item Adjust text or background color to avoid text color being too similar to the background color.
                \item If no improvement is necessary, leave ``\texttt{suggestions}'' as an empty string (``''); otherwise, briefly list the suggestion(s) in ``\texttt{suggestions}''.
              \end{itemize}
        \item Grade the response.
      \end{enumerate}

      \textbf{Output format (valid JSON)}
      \begin{verbatim}
```json
{
  "is_error": <boolean>,
  "error_message": "<string>",
  "screenshot_description": "<string>",
  "suggestions": "<string>"
}
```
      \end{verbatim}

      Return \textbf{only} this JSON object—no additional commentary, markdown, or code fences.
    \end{small}
  \end{tcolorbox}
  \caption{The prompt for generating the description and suggestions based on the website screenshot.}
  \label{fig:prompt_screenshot_desc}
\end{figure}

\begin{figure}
  \begin{tcolorbox}[colback=blue!5!white,colframe=blue!75!black]
    \begin{small}
    \textbf{Prompt:}\\
      You are tasked with evaluating the design of a webpage. Grade the webpage's appearance on a scale of 0 to 5 (5 being highest), considering the following criteria:

      \begin{itemize}[leftmargin=1.5em,itemsep=0.3em,topsep=0.3em]
        \item \textbf{Successful Rendering:} Are there any components in the page or is it completely blank? Does the webpage render correctly without visual errors? Are colors, fonts, and components displayed as specified?
        \item \textbf{Content Relevance:} Does the design align with the website’s purpose and user requirements? Are elements (e.g., search bars, report formats) logically placed and functional?
        \item \textbf{Layout Harmony:} Is the arrangement of components (text, images, buttons) balanced, intuitive, and clutter-free?
        \item \textbf{Modernness \& Beauty:} Does the design follow contemporary trends (e.g., minimalism, responsive layouts)? Are colors, typography, and visual hierarchy aesthetically pleasing?
      \end{itemize}

      \textbf{Grading Scale:}
      \begin{itemize}[leftmargin=1.5em,itemsep=0.3em,topsep=0.3em]
        \item \textbf{0 (Blank Page):} The screenshot is completely blank or does not contain any visible content. It may only have a background color or display an error message.
        \item \textbf{1 (Poor):} Major rendering issues (e.g., broken layouts, incorrect colors). Content is irrelevant or missing. Layout is chaotic. Design is outdated or visually unappealing.
        \item \textbf{2 (Below Average):} Partial rendering with noticeable errors. Content is partially relevant but poorly organized. Layout lacks consistency. Design is basic or uninspired.
        \item \textbf{3 (Average):} Mostly rendered correctly with minor flaws. Content is relevant but lacks polish. Layout is functional but unremarkable. Design is clean but lacks modern flair.
        \item \textbf{4 (Good):} Rendered well with no major errors. Content is relevant and logically organized. Layout is harmonious and user-friendly. Design is modern and visually appealing.
        \item \textbf{5 (Excellent):} Flawless rendering. Content is highly relevant, intuitive, and tailored to user needs. Layout is polished, responsive, and innovative. Design is cutting-edge, beautiful, and memorable.
      \end{itemize}

      \textbf{Task:}\\
      Review the provided screenshot(s) of the webpage. Provide a concise analysis of a few sentences and then assign a grade (0--5) based on your analysis. Highlight strengths, weaknesses, and how well the design adheres to the specifications.

      \textbf{Your Response Format}
      \begin{verbatim}
```json
{
  "analysis": "<string>",
  "grade": <int>
}
```
      \end{verbatim}

      \textbf{Your Response:}
    \end{small}
  \end{tcolorbox}
  \caption{Prompt for evaluating the visual quality of a webpage and generating an appearance score.}
  \label{fig:webpage_design_prompt}
\end{figure}

\begin{figure}
  \begin{tcolorbox}[colback=blue!5!white,colframe=blue!75!black]
    \begin{small}
      Based on the original website development instruction, you should identify \textbf{all the requirements} of the website generation and create a comprehensive instruction for a web-navigation GUI agent to test the generated website. The following is an example of triggering the GUI agent testing based on the original instruction:\\

      \textbf{\textit{Example}}\\

Original instruction: \\
Please implement a self-driving tour website that provides self-driving tour products and services. The website should have functionalities for browsing self-driving tour routes, booking self-driving tour hotels, and self-help self-driving tour packages. Users should be able to browse different types of self-driving tour routes, book hotels and packages, and query self-driving club information. The website should also provide search and filtering functions to help users quickly find the self-driving tour products they need. Define background as cream; define components with dark teal.\\

\texttt{<boltAction type="gui\_agent\_test">}\\
Verify cream background and dark-teal buttons. Browse different types of self-driving tour routes, book hotels and packages, and query self-driving club information. Search and filter for self-driving tour products.\\
\texttt{</boltAction>}\\

      The following is the original website development instruction:

\begin{verbatim}
<instruction>{instruction}</instruction>
\end{verbatim}

      Trigger the GUI agent testing based on the original instruction in a way similar to the example. \textbf{Do not generate additional comments.}
    \end{small}
  \end{tcolorbox}
  \caption{Prompt for generating a GUI-agent testing instruction from the original website specification.}
  \label{fig:gui_agent_trigger_prompt}
\end{figure}

\begin{figure}
  \begin{tcolorbox}[colback=blue!5!white,colframe=blue!75!black]
    \begin{small}
    \textbf{Prompt:}
      You are given a GUI-agent testing trajectory.\\

      \textbf{The GUI agent testing trajectory:}\\

      \textbf{GUI-agent Testing Instruction:}\\
      \texttt{\{gui\_instruction\}}\\

      \textbf{Trajectory:}\\
      \texttt{\{result\}}\\

      \textbf{Task}
      \begin{enumerate}[leftmargin=1.5em,itemsep=0.3em,topsep=0.3em]
        \item Examine the trajectory for any failed actions that indicate a problem in the website design.
        \item Decide whether the GUI-agent testing trajectory reveals any flaw in the website implementation.
              \begin{itemize}[leftmargin=1.5em,itemsep=0.2em,topsep=0.2em]
                \item If \textbf{yes}, set \texttt{"test\_passed": true}, and leave \texttt{"improvement\_suggestions"} empty.
                \item If \textbf{no}, set \texttt{"test\_passed": false}, and write a concise but thorough \texttt{"improvement\_suggestions"} that covers the suggested improvements targeting the problems revealed by the testing result.
              \end{itemize}
        \item Evaluate the results of the GUI-agent test run and assign \textbf{one integer grade from 1 to 5}:
              \begin{itemize}[leftmargin=1.5em,itemsep=0.2em,topsep=0.2em]
                \item 1: The vast majority of tested functions fail or behave incorrectly.
                \item 2: Many functions fail; only a few behave as expected.
                \item 3: About half of the functions work as expected; success is mixed.
                \item 4: Most functions work as expected; only minor issues remain.
                \item 5: All tested functions work exactly as expected; no issues observed.
              \end{itemize}
              Assign the grade to \texttt{"grade"}.
      \end{enumerate}

      \textbf{Output format (valid JSON)}
      \begin{verbatim}
```json
{
  "test_passed": <boolean>,
  "improvement_suggestions": "<string>",
  "grade": <int>
}
```
      \end{verbatim}

      You can first make a short analysis of two or three sentences, then output this JSON object.
    \end{small}
  \end{tcolorbox}
  \caption{Prompt for evaluating GUI-agent testing trajectories and providing improvement suggestions.}
  \label{fig:gui_agent_testing_prompt}
\end{figure}

\section{Examples of WebGen-Agent Trajectories}
\label{sec:trajectory_examples}

\begin{figure*}[t]
    \centering
    \includegraphics[width=1.0\textwidth]{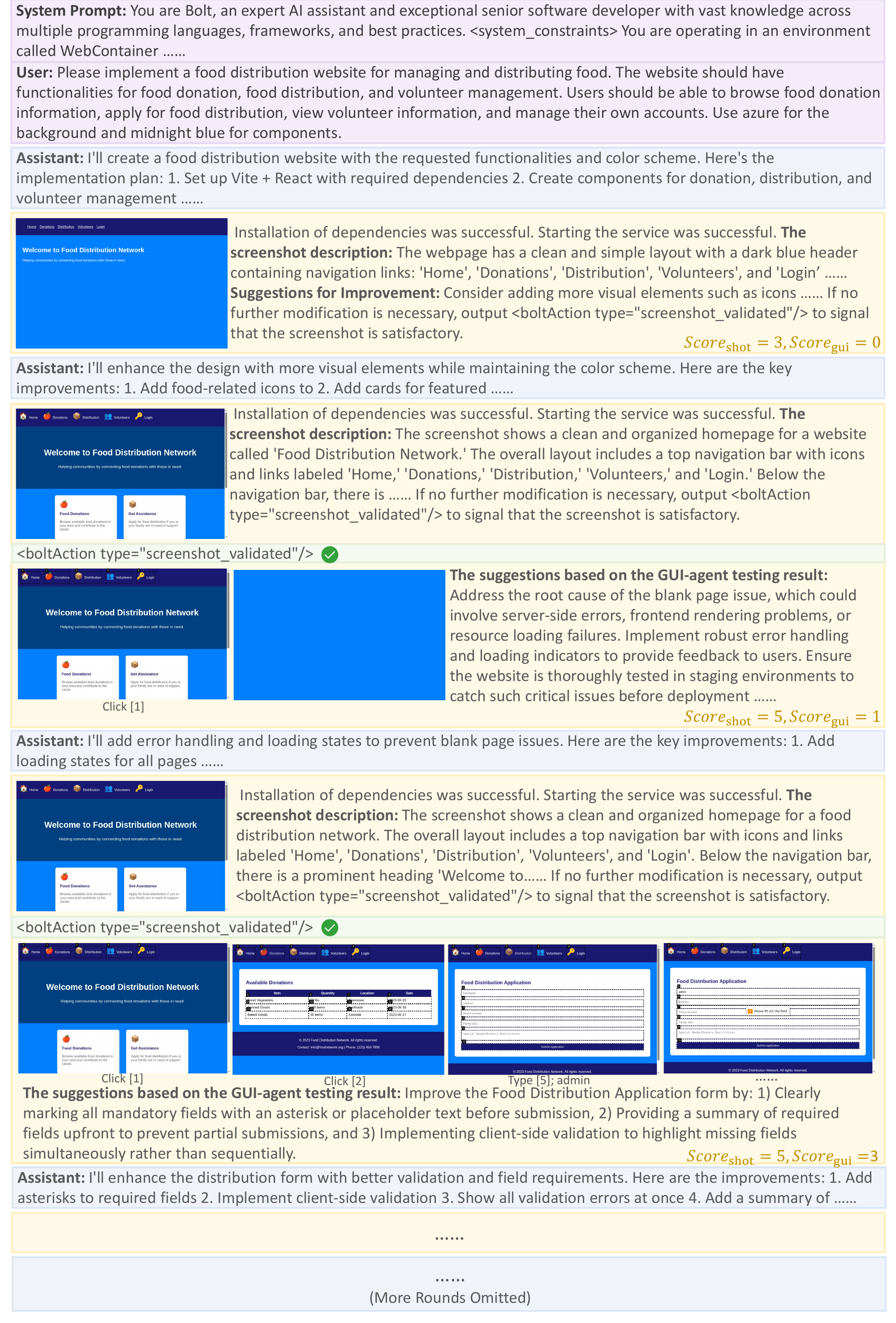}
    \caption{Example of a WebGen-Agent trajectory.}
    
\label{fig:pipeline_1}
\end{figure*}

\begin{figure*}[t]
    \centering
    \includegraphics[width=1.0\textwidth]{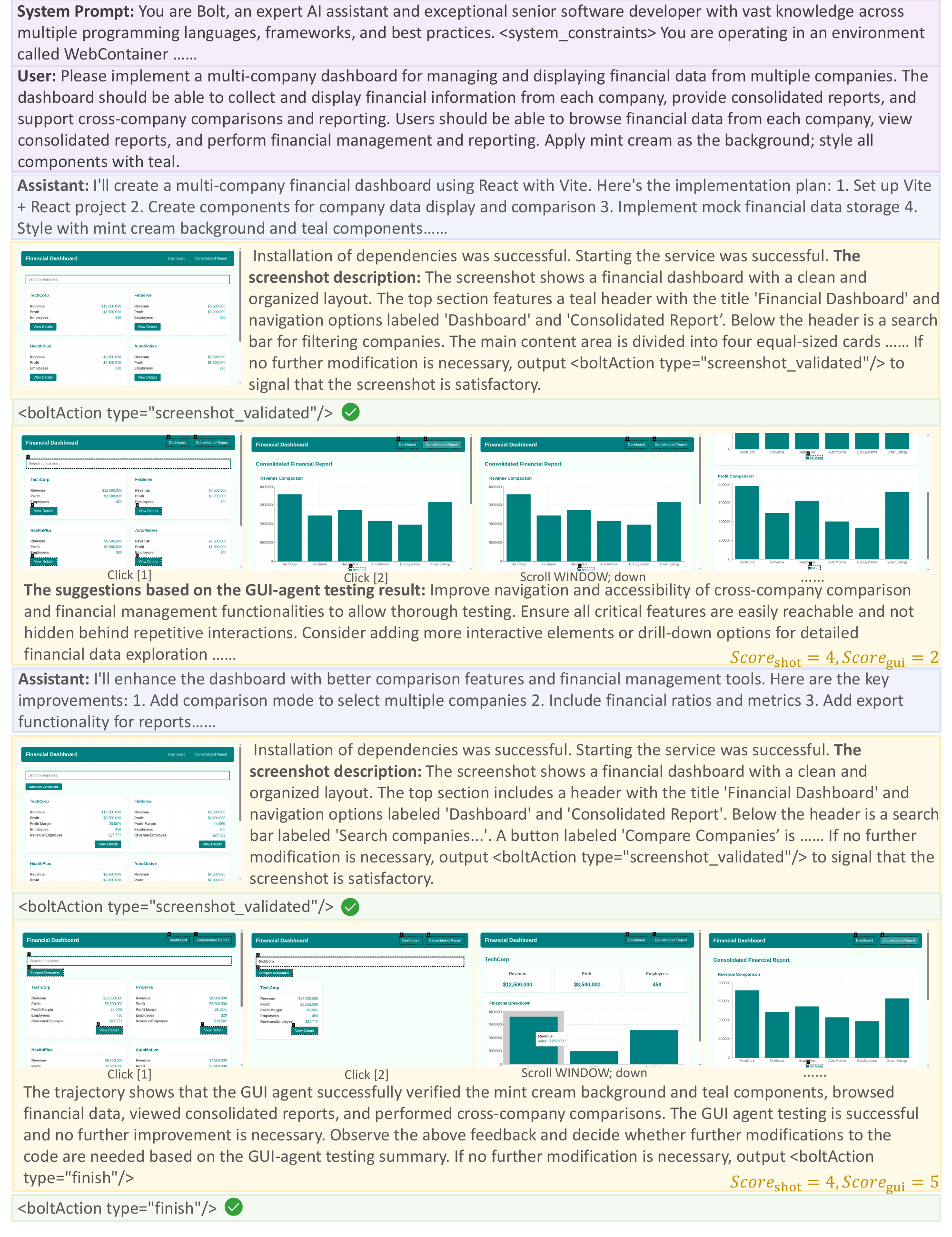}
    \caption{Example of a WebGen-Agent trajectory.}
    
\label{fig:pipeline_2}
\end{figure*}

\begin{figure*}[t]
    \centering
    \includegraphics[width=1.0\textwidth]{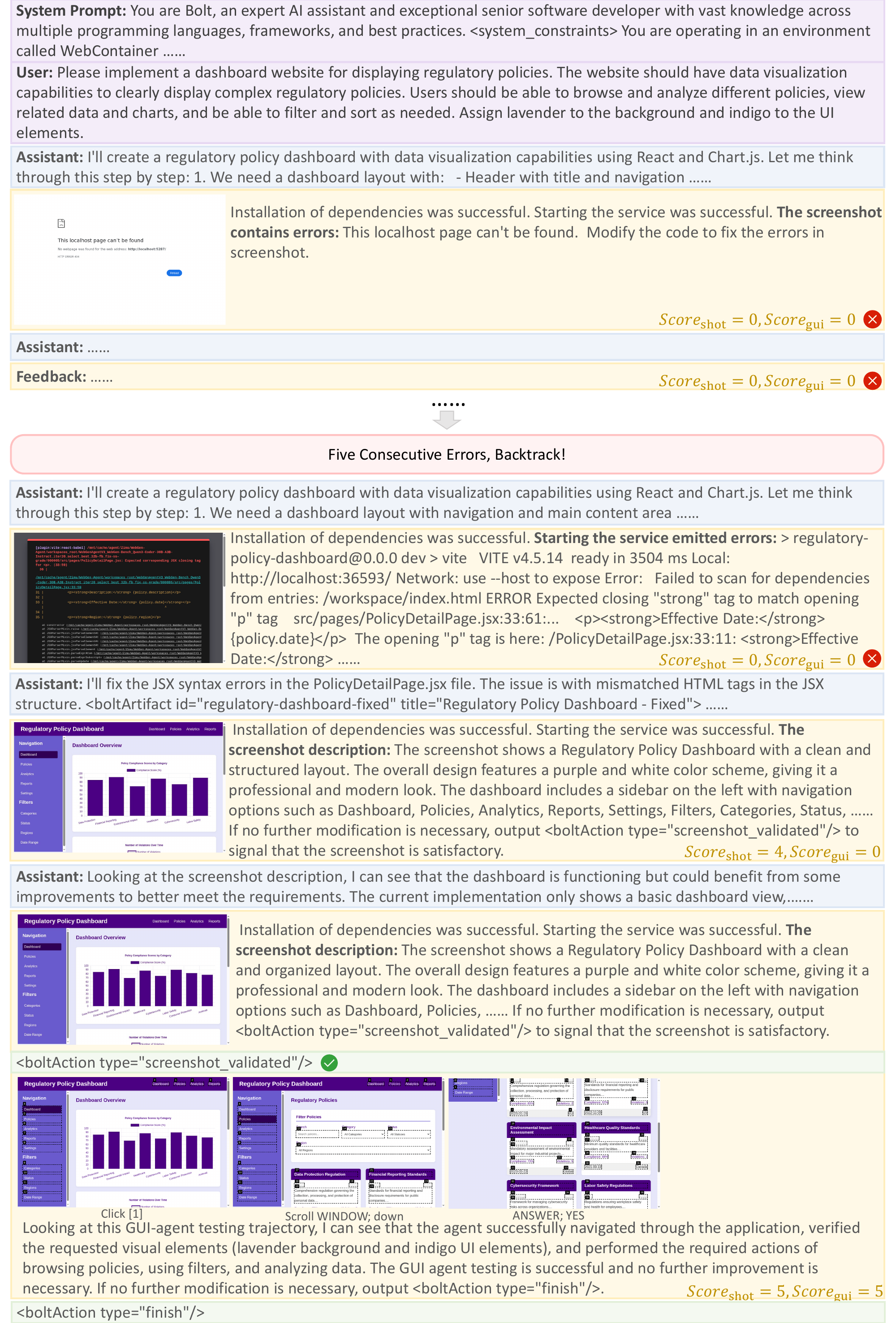}
    \caption{Example of a WebGen-Agent trajectory.}
    
\label{fig:pipeline_3}
\end{figure*}

To demonstrate the WebGen-Agent workflow in a straightforward way, we present three example trajectories in Fig.~\ref{fig:pipeline_1}, Fig.~\ref{fig:pipeline_2}, and Fig.~\ref{fig:pipeline_3}. As shown in these examples, WebGen-Agent iteratively improves the appearance and functionality of the generated website based on screenshot and GUI-agent testing feedback.

\section{Accuracy of Screenshot and GUI-agent Testing Scores}
\label{sec:score_acc}

To analyze the accuracy of the screenshot and GUI-agent testing scores given by the feedback VLM in the WebGen-Agent workflow, we evaluated the results of Claude-4-Sonnet, Qwen3-Coder-30B-A3B-Instruct, Qwen3-Coder-480B-A35B-Instruct, and DeepSeek-V3 as coding LLMs, with Qwen2.5-VL-32B-Instruct as the feedback VLM, as well as DeepSeek-V3 as the coding LLM and GPT-4o as the feedback VLM. We manually verified the accuracy of the screenshot and GUI-agent testing scores. Human annotators were provided with the score and the screenshot or GUI-agent trajectory at each step and asked to judge whether the score was accurate. If the score was inaccurate, they provided the correct score. The results are presented in Table~\ref{tab:score_accuracies}.

The accuracies of the screenshot scores across all experiments ranged from 93\% to 96\%, while the accuracies of the GUI-agent scores ranged from 89\% to 93\%. The standard errors of the screenshot scores range from 0.20 to 0.26, while the standard errors of the GUI-agent scores range from 0.31 to 0.44. This demonstrates that the scores are highly accurate, supporting the effectiveness of the WebGen-Agent workflow and the Step-GRPO training process. Compared with using Qwen2.5-VL-32B-Instruct, using GPT-4o as the feedback VLM only marginally improved the screenshot score accuracy from 94.8\% to 95.5\% and the GUI-agent score accuracy from 91.2\% to 92.2\%. This shows that Qwen2.5-VL-32B-Instruct is sufficient for the task while being significantly more cost-effective.

\begin{table}[t]
\fontsize{9}{10}\selectfont
\centering
\caption{Accuracy of the screenshot and GUI-agent scores using human annotation as ground truth. For every experiment we report the accuracy together with the standard error compared to human scores.}
\begin{tabularx}{\textwidth}
  {>{\raggedright\arraybackslash\hsize=0.5\hsize}X
   >{\raggedright\arraybackslash\hsize=1.6\hsize}X
   >{\raggedright\arraybackslash\hsize=1.2\hsize}X
   >{\centering\arraybackslash\hsize=0.4\hsize}X
   >{\centering\arraybackslash\hsize=0.5\hsize}X}
\toprule
\textbf{Score Type} & \textbf{Coding LLM} & \textbf{Feedback VLM} &
\textbf{Accuracy (\%)} & \textbf{Std.\ Error} \\
\midrule\midrule
\multirow{5}{*}{Screenshot}
  & Claude-4-Sonnet                 & Qwen2.5-VL-32B-Inst. & 93.6 & 0.25 \\
  & Qwen3-Coder-30B-A3B-Inst.       & Qwen2.5-VL-32B-Inst. & 93.9 & 0.26 \\
  & Qwen3-Coder-480B-A35B-Inst. & Qwen2.5-VL-32B-Inst. & 95.6 & 0.20 \\
  & DeepSeek-V3                     & Qwen2.5-VL-32B-Inst. & 94.8 & 0.22 \\
  & DeepSeek-V3                     & GPT-4o               & 95.5 & 0.20 \\
\midrule
\multirow{5}{*}{GUI agent}
  & Claude-4-Sonnet                 & Qwen2.5-VL-32B-Inst. & 90.1 & 0.31 \\
  & Qwen3-Coder-30B-A3B-Inst.       & Qwen2.5-VL-32B-Inst. & 91.4 & 0.44 \\
  & Qwen3-Coder-480B-A35B-Inst.     & Qwen2.5-VL-32B-Inst. & 89.6 & 0.41 \\
  & DeepSeek-V3                     & Qwen2.5-VL-32B-Inst. & 91.2 & 0.36 \\
  & DeepSeek-V3            & GPT-4o      & 92.2 & 0.33 \\
\bottomrule
\end{tabularx}
\label{tab:score_accuracies}
\end{table}

\section{Analysis of the Comprehensiveness of GUI-agent Testing Instructions}
\label{sec:gui_instruction_comprehensiveness}

To analyze the comprehensiveness of the GUI-agent testing instructions generated by the agent, we manually evaluated the instructions from the experiment runs using Claude-4-Sonnet, Qwen3-Coder-30B-A3B-Instruct, Qwen3-Coder-480B-A35B-Instruct, and DeepSeek-V3. We graded each GUI-agent instruction on a 1–5 scale, determined by how completely the instruction translates each website requirement into concrete GUI-agent checks. The grading guidelines are presented in Fig.~\ref{fig:gui_instruction_grading}.

As shown in Tab.~\ref{tab:gui_instruction_scores}, 77.2\% of the GUI-agent testing instructions across the four models receive a score of 5 (Complete, $\approx$ 100\% of requirements). Instructions with a score of 4 or higher (High, 75–90\%) account for 98.3\% of the total, while only 1.7\% receive a score of 3 (Moderate, 50–75\%); none score below 3. These results indicate that the GUI-agent instructions comprehensively cover most of the website requirements.

\begin{table}[t]
\fontsize{9}{10}\selectfont
\centering
\caption{Distribution (\%) of human scores regarding the comprehensiveness of the GUI-agent testing instructions and the resulting average score. The definition of the scores are presented in Fig.~\ref{fig:gui_instruction_grading}. The scores range from 1 to 5.}
\begin{tabularx}{\textwidth}
  {>{\raggedright\arraybackslash\hsize=1.6\hsize}X
   >{\centering\arraybackslash\hsize=0.35\hsize}X
   >{\centering\arraybackslash\hsize=0.35\hsize}X
   >{\centering\arraybackslash\hsize=0.35\hsize}X
   >{\centering\arraybackslash\hsize=0.35\hsize}X
   >{\centering\arraybackslash\hsize=0.35\hsize}X
   >{\centering\arraybackslash\hsize=0.55\hsize}X}
\toprule
\textbf{Model} & \textbf{5} & \textbf{4} & \textbf{3} & \textbf{2} & \textbf{1} & \textbf{Avg.\ Score} \\
\midrule\midrule
Claude-4-Sonnet              & 84.2 & 13.9 & 2.0 & 0.0 & 0.0 & 4.82 \\
DeepSeek-V3                  & 73.3 & 24.8 & 2.0 & 0.0 & 0.0 & 4.71 \\
Qwen3-Coder-30B-A3B-Inst.    & 75.2 & 23.8 & 1.0 & 0.0 & 0.0 & 4.74 \\
Qwen3-Coder-480B-A35B-Inst.  & 76.2 & 21.8 & 2.0 & 0.0 & 0.0 & 4.74 \\
\midrule
Total                        & 77.2 & 21.0 & 1.7 & 0.0 & 0.0 & 4.75 \\
\bottomrule
\end{tabularx}
\label{tab:gui_instruction_scores}
\end{table}

\begin{figure}
  \begin{tcolorbox}[colback=blue!5!white,colframe=blue!75!black]
    \begin{small}

      \textbf{GUI-agent Instruction Evaluation Guidelines:}\\
Score the instruction 1 – 5, where 5 = best. The dominant criterion is comprehensiveness: how completely the instruction translates every website requirement into concrete GUI-agent checks.\\

\textbf{Grading Scale:}
      \begin{itemize}[leftmargin=1.5em,itemsep=0.3em,topsep=0.3em]
        \item \textbf{1 (Minimal, $<$ 25 \%):} The instruction overlooks most of the stated requirements.
        \item \textbf{2 (Low, 25 – 50 \%):} Only some primary requirements are mentioned; many important items are absent.
        \item \textbf{3 (Moderate, 50 – 75 \%):} Core functionalities are covered, but several secondary features or style rules are skipped.
        \item \textbf{4 (High, 75 – 90 \%):} All major functional requirements plus most visual or secondary ones are included; only a few minor details are missing.
        \item \textbf{5 (Complete, $\approx$ 100 \% of requirements):} Every requirement is turned into checks. Nothing significant is left out.
      \end{itemize}

    \end{small}
  \end{tcolorbox}
  \caption{Grading guidelines for manually evaluating GUI-agent testing instructions}
  \label{fig:gui_instruction_grading}
\end{figure}

\section{Categorical Results}
\label{sec:categorical_results}

Tab.~\ref{tab:categorical_results} shows the categorical results of WebGen-Agent with various proprietary and open-source models on WebGen-Bench. As shown in the table, WebGen-Agent consistently achieves superior performance across all instruction and test-case categories compared to other code agent systems. For both the 7B and 8B models, Step-GRPO improves performance in most categories compared to the original instruct model and the SFT model. This demonstrates the effectiveness of the WebGen-Agent workflow and the Step-GRPO training process, which incorporates screenshots and GUI-agent feedback.

\begin{table}[t]\fontsize{8.5}{10}\selectfont
\centering
\caption{Categorical results of WebGen-Agent with various proprietary and open-source models on WebGen-Bench~\citep{lu2025webgenbenchevaluatingllmsgenerating}, compared with other code agent systems. The highest score of each column is marked in \textbf{bold}.}
\begin{tabularx}{\textwidth}
  {>{\raggedright\arraybackslash\hsize=2.8\hsize}X
   >{\centering\arraybackslash\hsize=0.6\hsize}X
   >{\centering\arraybackslash\hsize=0.6\hsize}X
   >{\centering\arraybackslash\hsize=0.7\hsize}X
   >{\centering\arraybackslash\hsize=0.6\hsize}X
   >{\centering\arraybackslash\hsize=0.6\hsize}X
   >{\centering\arraybackslash\hsize=0.6\hsize}X}
\toprule
\multirow{2}{\hsize}{\textbf{Test Name}} &
\multicolumn{3}{c}{\textbf{Instruction Categories}} &
\multicolumn{3}{c}{\textbf{Test-case Categories}} \\
\cmidrule(r){2-4}\cmidrule(r){5-7}
 & \textbf{Content Presentation} & \textbf{User Interaction} &
   \textbf{Data Management} & \textbf{Functional Testing} &
   \textbf{Data-Display Testing} & \textbf{Design-Validation} \\
\midrule\midrule
\multicolumn{7}{c}{\textbf{OpenHands}}\\ \midrule
Claude-3.5-Sonnet & 32.8 & 18.4 & 18.4 & 12.4 & 33.9 & 32.0 \\
DeepSeek-R1       & 16.4 &  8.9 &  5.9 &  5.0 &  9.9 & 25.0 \\
DeepSeek-V3       & 12.6 &  7.3 &  8.4 &  3.8 &  8.1 & 25.0 \\
\midrule
\multicolumn{7}{c}{\textbf{Aider}}\\ \midrule
Claude-3.5-Sonnet & 31.9 & 21.1 & 16.6 & 14.9 & 30.1 & 34.0 \\
DeepSeek-R1       & 39.1 & 28.6 & 13.4 & 17.6 & 35.2 & 44.3 \\
DeepSeek-V3       & 17.8 & 12.8 & 12.5 &  9.7 & 19.1 & 18.4 \\
\midrule\midrule
\multicolumn{7}{c}{\textbf{Bolt.diy}}\\ \midrule
Claude-3.5-Sonnet & 35.6 & 21.2 & 26.2 & 17.1 & 26.3 & 52.0 \\
DeepSeek-R1       & 43.7 & 20.6 & 24.7 & 21.1 & 29.3 & 44.3 \\
DeepSeek-V3       & 37.1 & 16.6 & 11.2 & 10.5 & 28.2 & 38.1 \\
GPT-4o            & 26.4 &  5.9 & 11.2 &  4.7 & 19.6 & 24.6 \\
o3-mini           & 28.7 & 17.7 & 13.4 & 11.4 & 25.5 & 33.6 \\
Qwen2.5-Coder-32B & 17.5 &  6.9 &  5.9 &  1.9 & 14.5 & 23.0 \\
Qwen2.5-72B-Inst. & 28.2 & 10.1 &  5.6 &  5.8 & 21.0 & 25.4 \\
WebGen-LM-7B      & 27.9 & 23.8 & 38.1 & 22.0 & 27.7 & 47.5 \\
WebGen-LM-14B     & 30.2 & 27.8 & 31.6 & 23.6 & 26.9 & 49.2 \\
WebGen-LM-32B     & 46.6 & 33.2 & 38.8 & 29.1 & 43.0 & 56.1 \\
\midrule\midrule
\multicolumn{7}{c}{\textbf{WebGen-Agent}}\\ \midrule
\multicolumn{7}{c}{\textbf{Proprietary Models}}\\ \midrule
Claude-3.5-Sonnet & 57.8 & 48.7 & 51.9 & 38.5 & 60.5 & 76.2 \\
DeepSeek-R1       & 57.8 & 44.2 & 38.1 & 35.0 & 53.8 & 66.8 \\
DeepSeek-V3       & 58.0 & 53.2 & 45.6 & 40.9 & 61.0 & 72.5 \\
o3                & 59.2 & 46.6 & 53.4 & 43.7 & 55.1 & 68.9 \\
Claude-4-Sonnet   & \textbf{68.7} & 51.8 & 52.5 & \textbf{44.0} & 69.4 & 71.7 \\
Gemini-2.5-Pro    & 60.3 & 48.2 & 45.6 & 37.9 & 60.2 & 72.5 \\
Qwen3-Coder-480B-A35B-Inst. &64.7	&\textbf{55.8}	&\textbf{55.9}	&43.2	&\textbf{71.2}	&\textbf{79.9}\\
\midrule
\multicolumn{7}{c}{\textbf{Open-Source Models (30B--72B)}}\\ \midrule
Qwen2.5-Coder-32B-Inst.   & 35.6 & 28.8 & 34.4 & 20.9 & 32.3 & 62.3 \\
Qwen3-Coder-30B-A3B-Inst. & \textbf{55.2} & \textbf{54.3} & \textbf{47.2} & \textbf{39.1} & \textbf{62.1} & \textbf{76.6} \\
Qwen2.5-72B-Instruct      & 43.4	&30.4	&38.8	&23.0	&39.8	&66.0 \\
\midrule
\multicolumn{7}{c}{\textbf{Open-Source Models (7B--8B)}}\\ \midrule
Qwen2.5-Coder-7B-Inst.                     & 20.7 &  8.6 & 10.9 &  7.4 & 15.9 & 21.3 \\
WebGenAgent-LM-7B-SFT                & \textbf{53.4} & 33.5 & 33.8 & 23.5 & 48.4 & 67.6 \\
WebGenAgent-LM-7B-Step-GRPO           & 51.1 & \textbf{41.1} & \textbf{47.8} & \textbf{30.7} & \textbf{56.7} & \textbf{69.3} \\
\midrule
Qwen3-8B                                   & 37.4 & 34.3 & 30.0 & 26.8 & 34.1 & 54.1 \\
WebGenAgent-LM-8B-SFT                               & 41.7 & 34.2 & \textbf{43.8} & 26.8 & 43.8 & 63.1 \\
WebGenAgent-LM-8B-Step-GRPO                         & \textbf{52.0}	&\textbf{38.8}	&43.1	&\textbf{30.2}	&\textbf{51.1}	&\textbf{68.4} \\
\bottomrule
\end{tabularx}
\label{tab:categorical_results}
\end{table}

\section{Analysis of Maximum Iteration Numbers}
\label{sec:maximum_iteration_number}

To analyze the effect of the maximum iteration number parameter on the performance of WebGen-Agent, we test the accuracy, appearance score, and the percentage of samples that exceed the maximum iteration limit (exceed rate) at different maximum iteration numbers. The coding LLM used is DeepSeek-V3.

\begin{figure*}[t]
  \centering
  \begin{subfigure}[t]{0.49\textwidth}
    \centering
    \includegraphics[width=\linewidth]{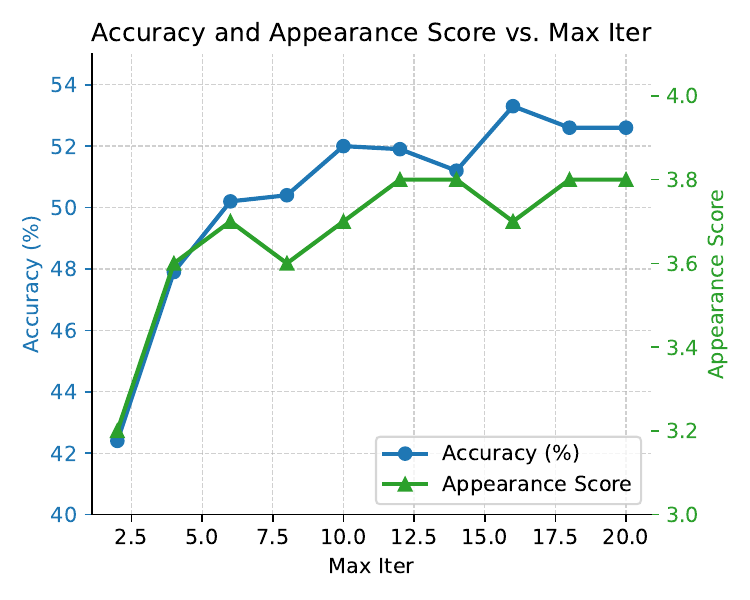}
    \caption{Accuracy (\%) and Appearance Score as a function of the maximum number of iterations.\label{fig:acc_app}}
  \end{subfigure}
  \hfill
  \begin{subfigure}[t]{0.49\textwidth}
    \centering
    \includegraphics[width=\linewidth]{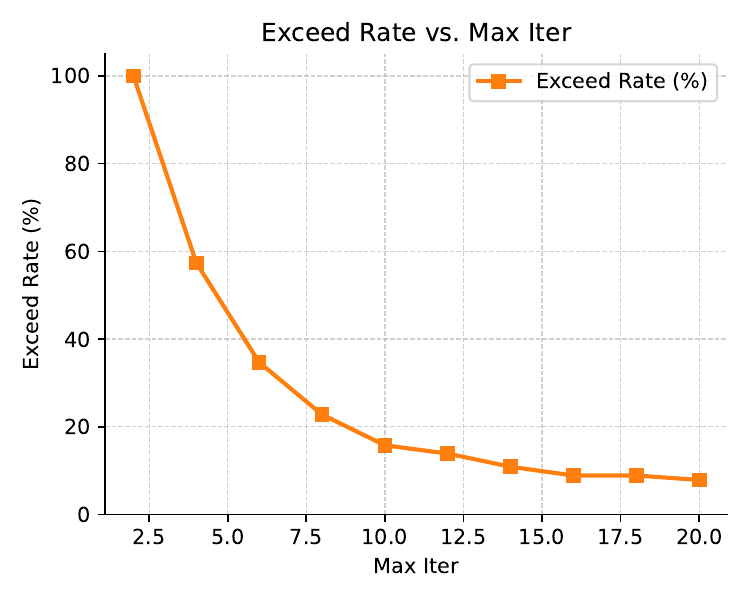}
    \caption{Exceed Rate (\%) versus the maximum number of iterations.\label{fig:exceed}}
  \end{subfigure}
  \caption{Effect of the maximum iteration number hyper-parameter on different performance metrics.}
  \label{fig:max_iter_metrics}
\end{figure*}

As shown in Fig.~\ref{fig:max_iter_metrics} and Tab.~\ref{tab:max_iter_ablation}, the accuracy and appearance score show a rising trend as the maximum iteration number increases, while the exceed rate continuously decreases. When the maximum iteration number is between 14 and 20, the accuracy, appearance score, and exceed rate all begin to converge. This is because most samples finish before reaching the iteration limit, as reflected by the exceed rate, and the impact of the maximum iteration number on performance diminishes.

\begin{table}[t]\fontsize{9}{10}\selectfont
\centering
\caption{Influence of the maximum number of iterations on WebGen-Agent performance.}
\begin{tabularx}{\textwidth}
  {>{\raggedright\arraybackslash\hsize=1.4\hsize}X
   >{\centering\arraybackslash\hsize=0.35\hsize}X
   >{\centering\arraybackslash\hsize=0.35\hsize}X
   >{\centering\arraybackslash\hsize=0.35\hsize}X
   >{\centering\arraybackslash\hsize=0.35\hsize}X
   >{\centering\arraybackslash\hsize=0.35\hsize}X
   >{\centering\arraybackslash\hsize=0.35\hsize}X
   >{\centering\arraybackslash\hsize=0.35\hsize}X
   >{\centering\arraybackslash\hsize=0.35\hsize}X
   >{\centering\arraybackslash\hsize=0.35\hsize}X
   >{\centering\arraybackslash\hsize=0.35\hsize}X}
\toprule
\textbf{Metric} & \textbf{2} & \textbf{4} & \textbf{6} & \textbf{8} & \textbf{10} & \textbf{12} & \textbf{14} & \textbf{16} & \textbf{18} & \textbf{20} \\
\midrule
Accuracy (\%) & 42.4 & 47.9 & 50.2 & 50.4 & 52.0 & 51.9 & 51.2 & 53.3 & 52.6 & 52.6 \\
Appearance Score & 3.2 & 3.6 & 3.7 & 3.6 & 3.7 & 3.8 & 3.8 & 3.7 & 3.8 & 3.8 \\
Exceed Rate (\%) & 100.0 & 57.4 & 34.7 & 22.8 & 15.8 & 13.9 & 10.9 & 8.9 & 8.9 & 7.9 \\
\bottomrule
\end{tabularx}
\label{tab:max_iter_ablation}
\end{table}

\section{Qualitative Analysis of Supervised Finetuning and Step-GRPO}
\label{sec:sft_step_grpo_qualitative_analysis}

\begin{figure*}[t]
    \centering
    \includegraphics[width=1.0\textwidth]{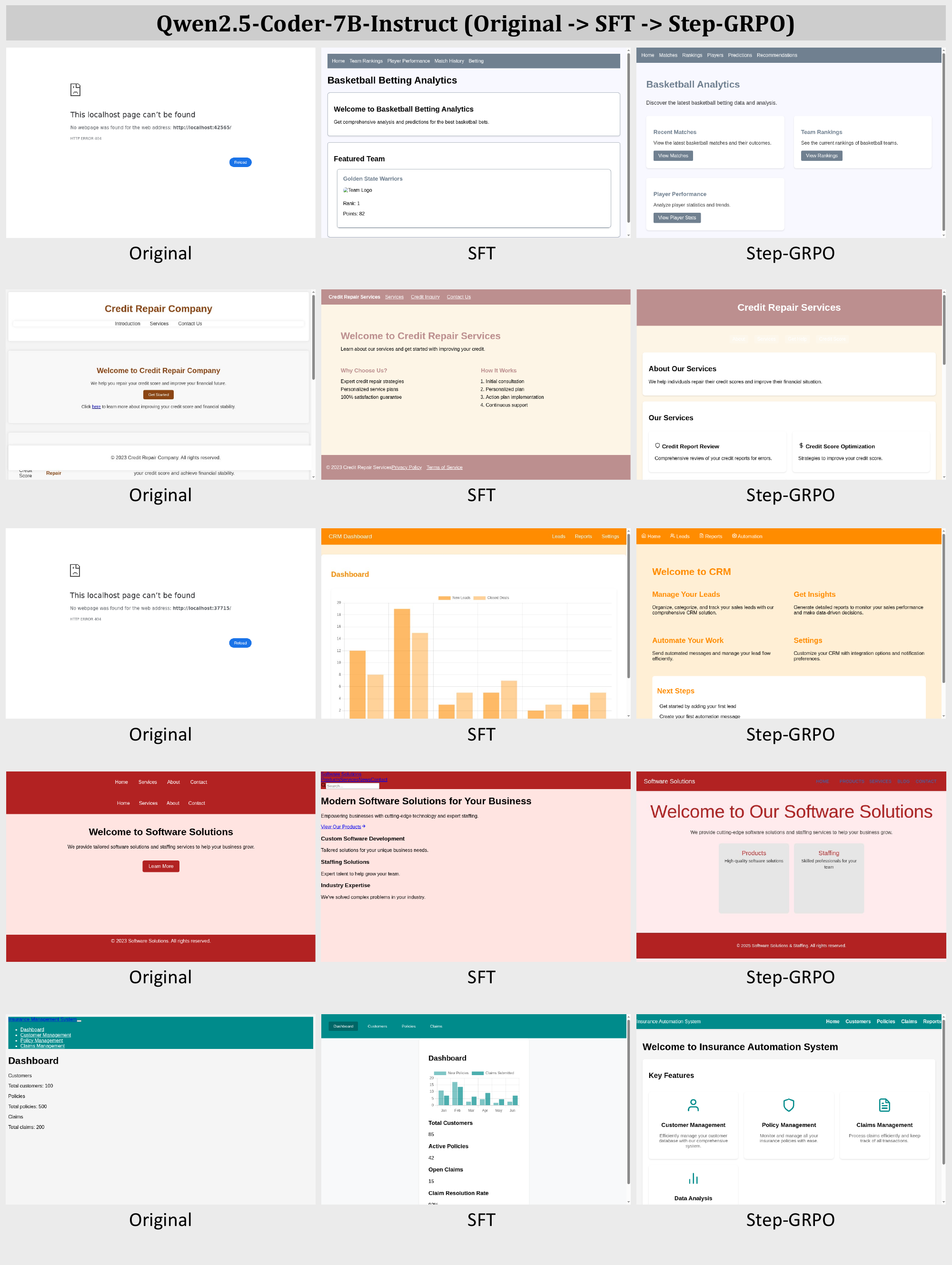}
    \caption{Screenshots of websites created by Qwen2.5-Coder-7B-Instruct, WebGenAgent-LM-7B-SFT, and WebGenAgent-LM-7B-Step-GRPO.}
    
\label{fig:shots_qwen2_5-coder-7B_1}
\end{figure*}

\begin{figure*}[t]
    \centering
    \includegraphics[width=1.0\textwidth]{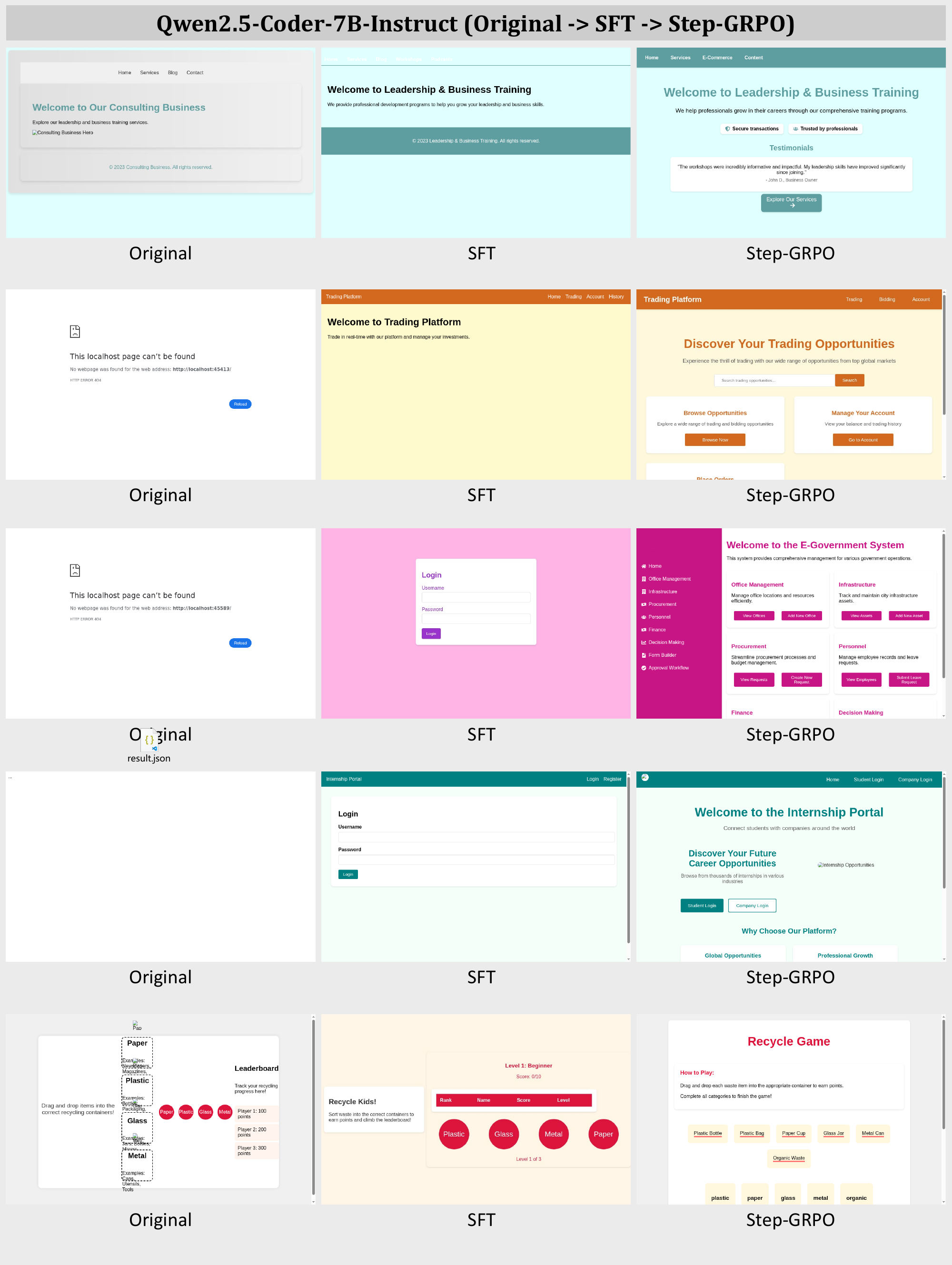}
    \caption{Screenshots of websites created by Qwen2.5-Coder-7B-Instruct, WebGenAgent-LM-7B-SFT, and WebGenAgent-LM-7B-Step-GRPO.}
    
\label{fig:shots_qwen2_5-coder-7B_2}
\end{figure*}

\begin{figure*}[t]
    \centering
    \includegraphics[width=1.0\textwidth]{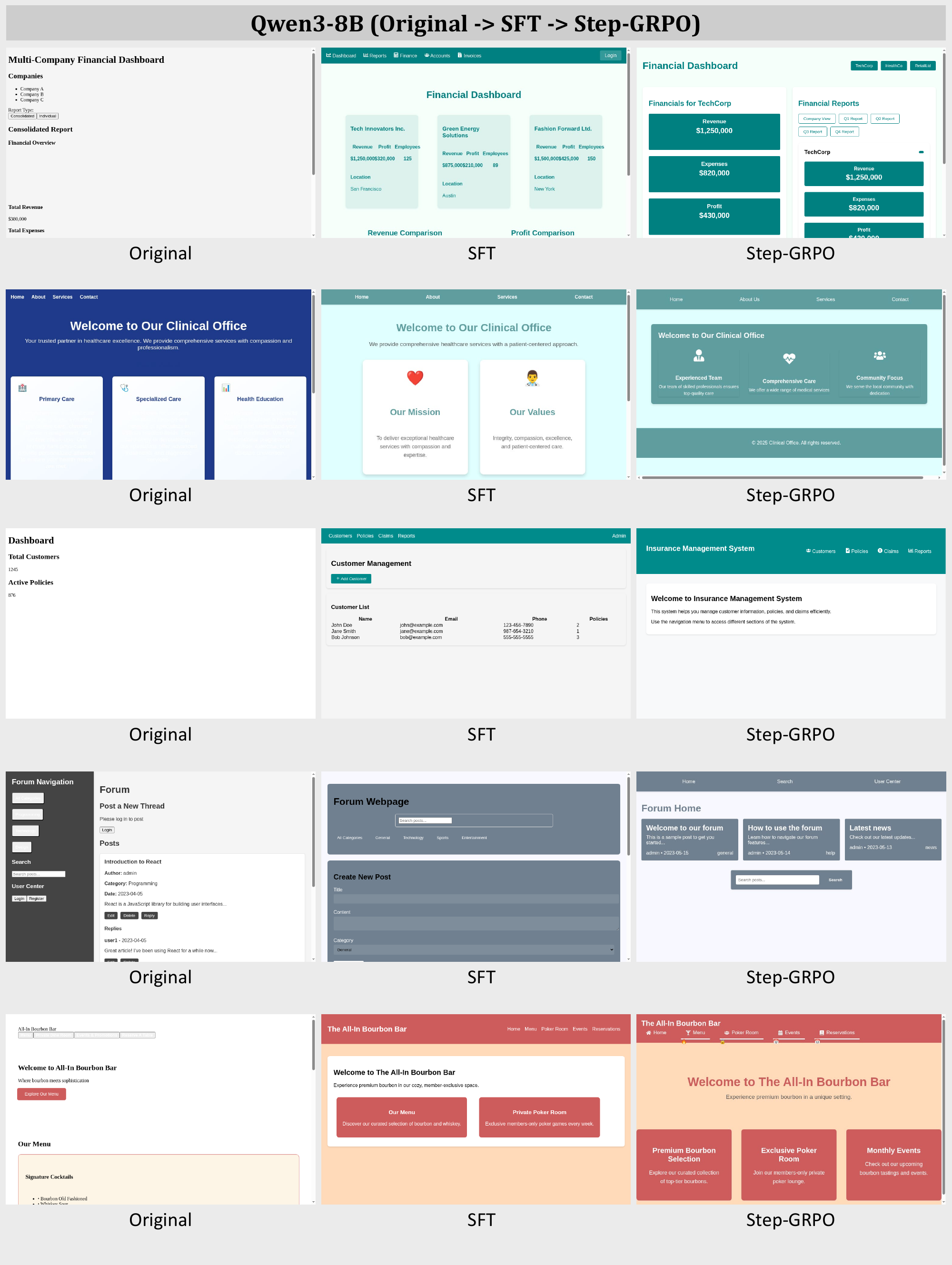}
    \caption{Screenshots of websites created by Qwen3-8B, WebGenAgent-LM-8B-SFT, and WebGenAgent-LM-8B-Step-GRPO.}
    
\label{fig:shots_qwen3-8B_1}
\end{figure*}

\begin{figure*}[t]
    \centering
    \includegraphics[width=1.0\textwidth]{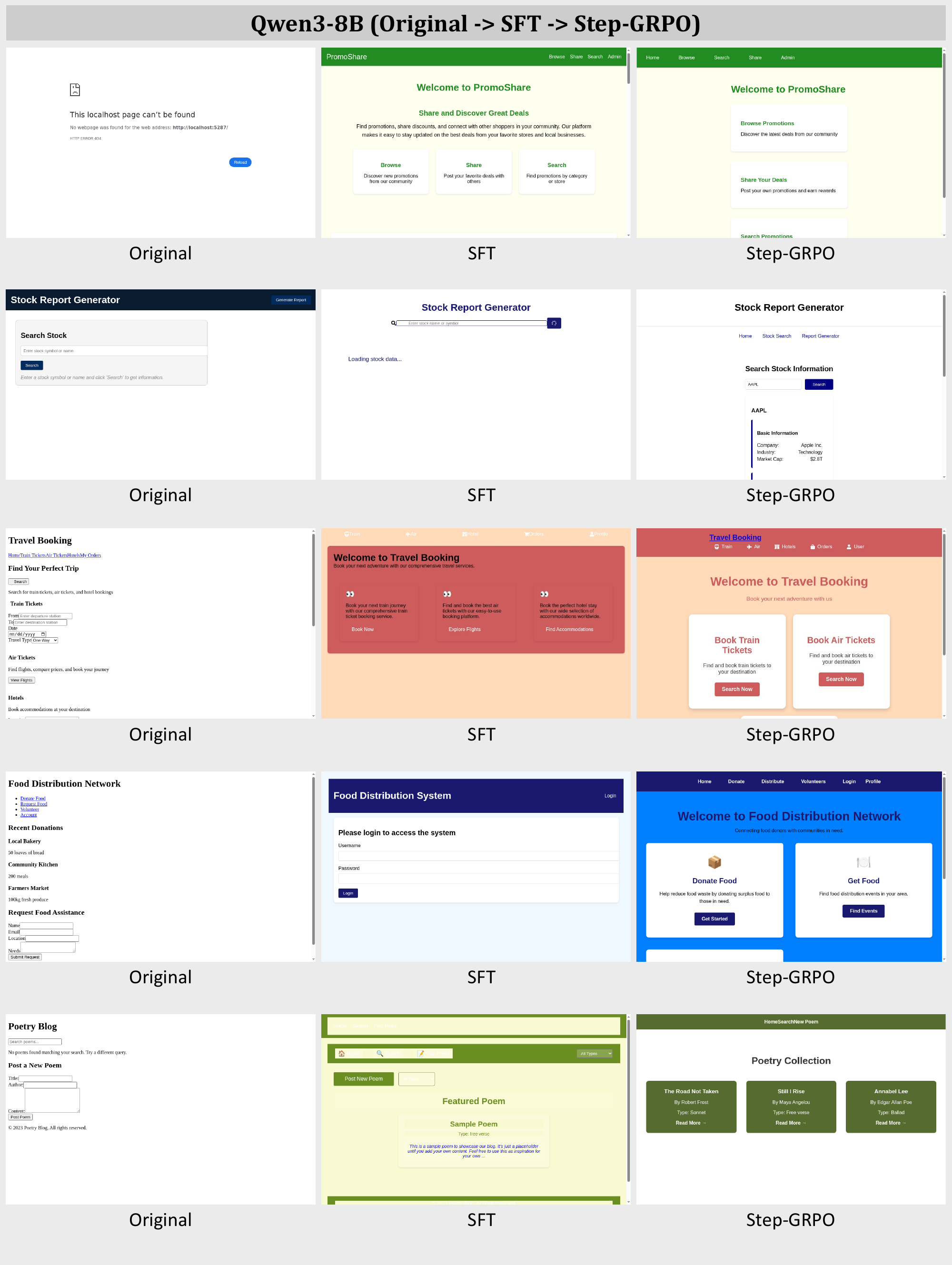}
    \caption{Screenshots of websites created by Qwen3-8B, WebGenAgent-LM-8B-SFT, and WebGenAgent-LM-8B-Step-GRPO.}
    
\label{fig:shots_qwen3-8B_2}
\end{figure*}

To provide a qualitative analysis of the effects of supervised fine-tuning and Step-GRPO with screenshot and GUI-agent feedback, we present examples of websites generated by Qwen2.5-Coder-7B-Instruct, WebGenAgent-LM-7B-SFT, and WebGenAgent-LM-7B-Step-GRPO in Figs.~\ref{fig:shots_qwen2_5-coder-7B_1} and~\ref{fig:shots_qwen2_5-coder-7B_2}. We also include examples of websites generated by Qwen3-8B, WebGenAgent-LM-8B-SFT, and WebGenAgent-LM-8B-Step-GRPO in Figs.~\ref{fig:shots_qwen3-8B_1} and\ref{fig:shots_qwen3-8B_2}. As demonstrated in the examples, supervised fine-tuning greatly reduces the models' tendency to generate erroneous or malformed websites and improves their ability to follow the appearance requirements specified in the instructions. Step-GRPO further refines the aesthetics and harmony of the generated websites.

\section{Qualitative Analysis of the WebGen-Agent Workflow}
\label{sec:workflow_qualitative_analysis}

\begin{figure*}[t]
    \centering
    \includegraphics[width=1.0\textwidth]{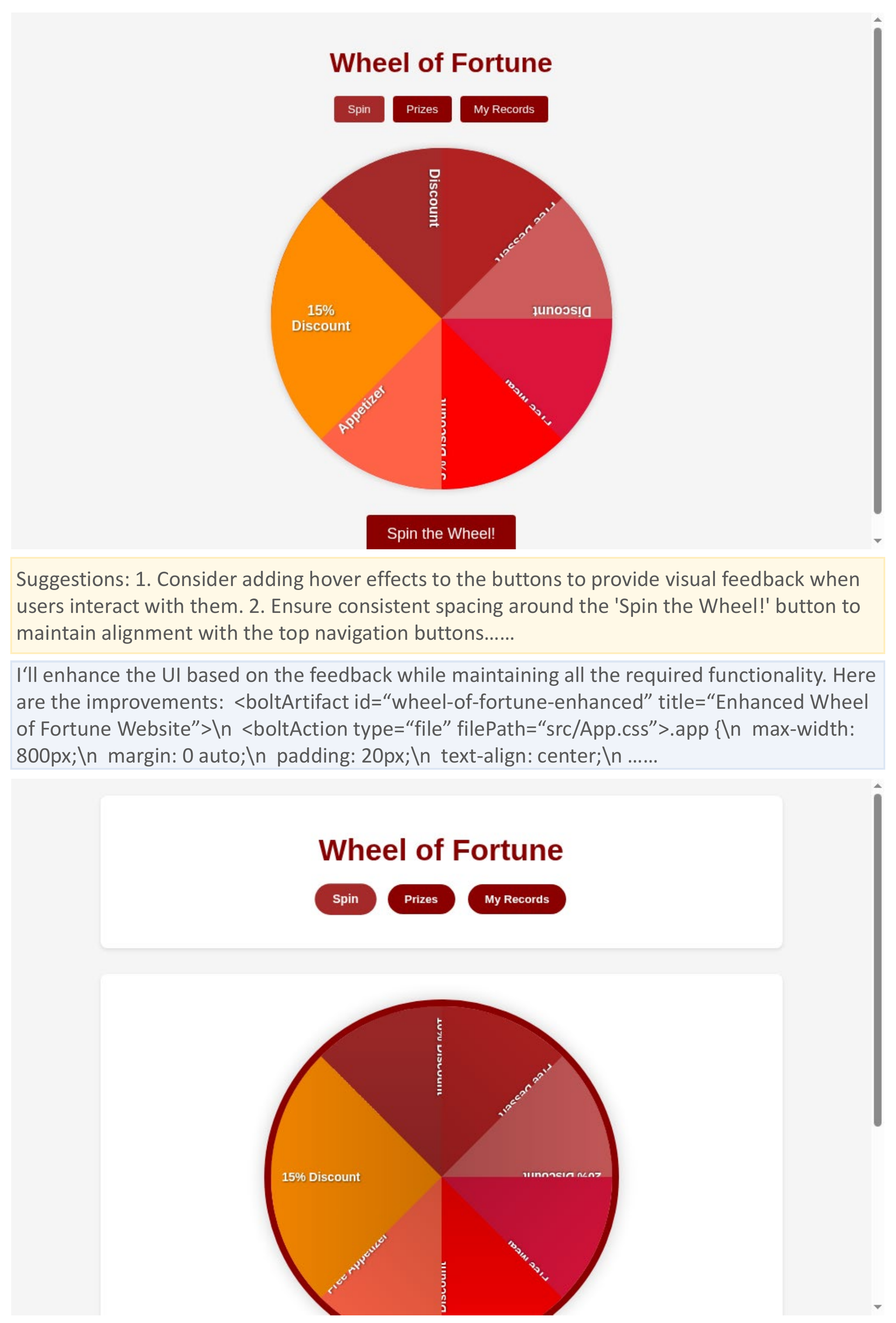}
    \caption{Example of a step in a WebGen-Agent trajectory where the agent improves the website's appearance based on screenshot feedback. The step is simplified due to space constraints.}
    
\label{fig:screenshot_improve_1}
\end{figure*}

\begin{figure*}[t]
    \centering
    \includegraphics[width=1.0\textwidth]{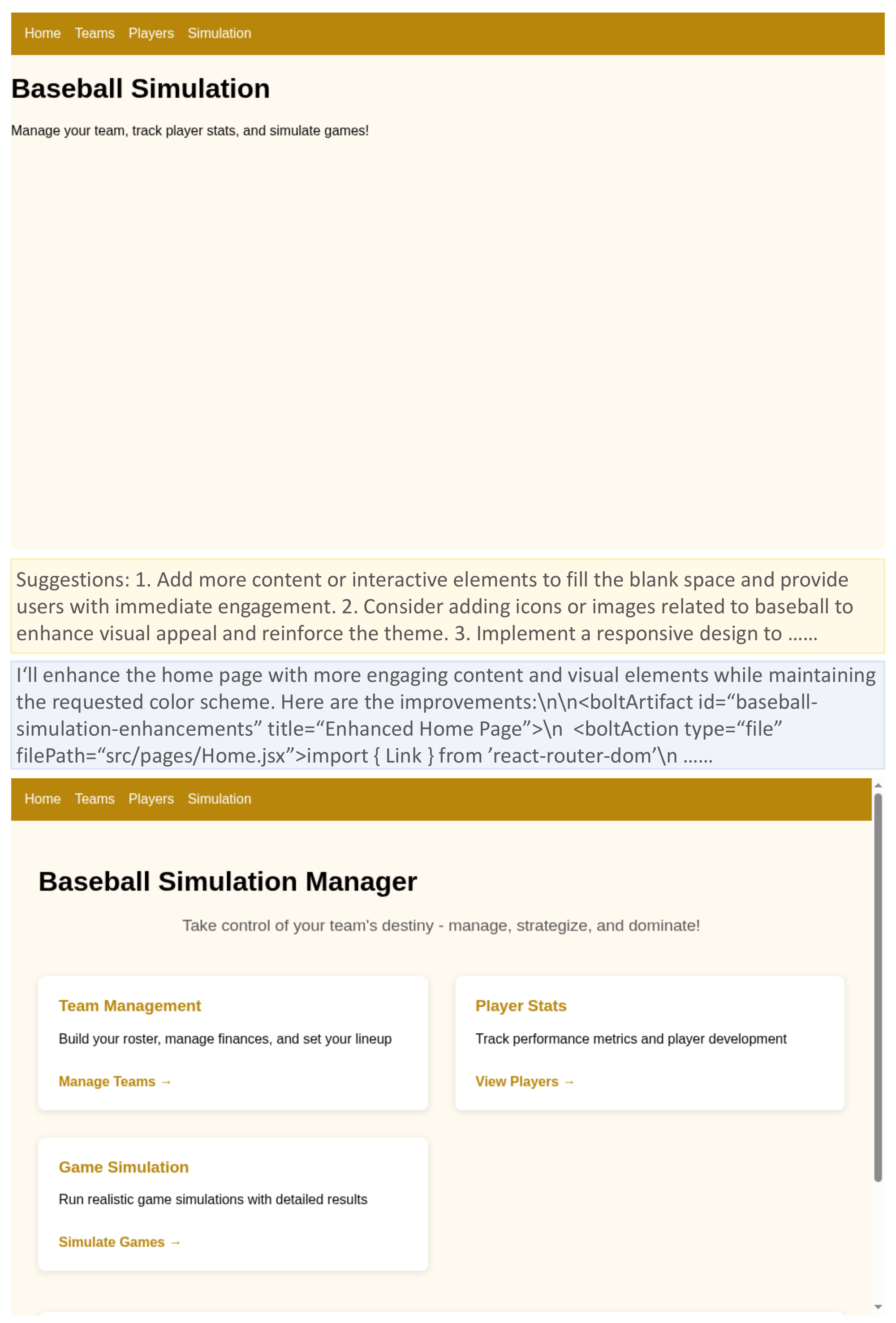}
    \caption{Example of a step in a WebGen-Agent trajectory where the agent improves the website's appearance based on screenshot feedback. The step is simplified due to space constraints.}
    
\label{fig:screenshot_improve_2}
\end{figure*}

\begin{figure*}[t]
    \centering
    \includegraphics[width=1.0\textwidth]{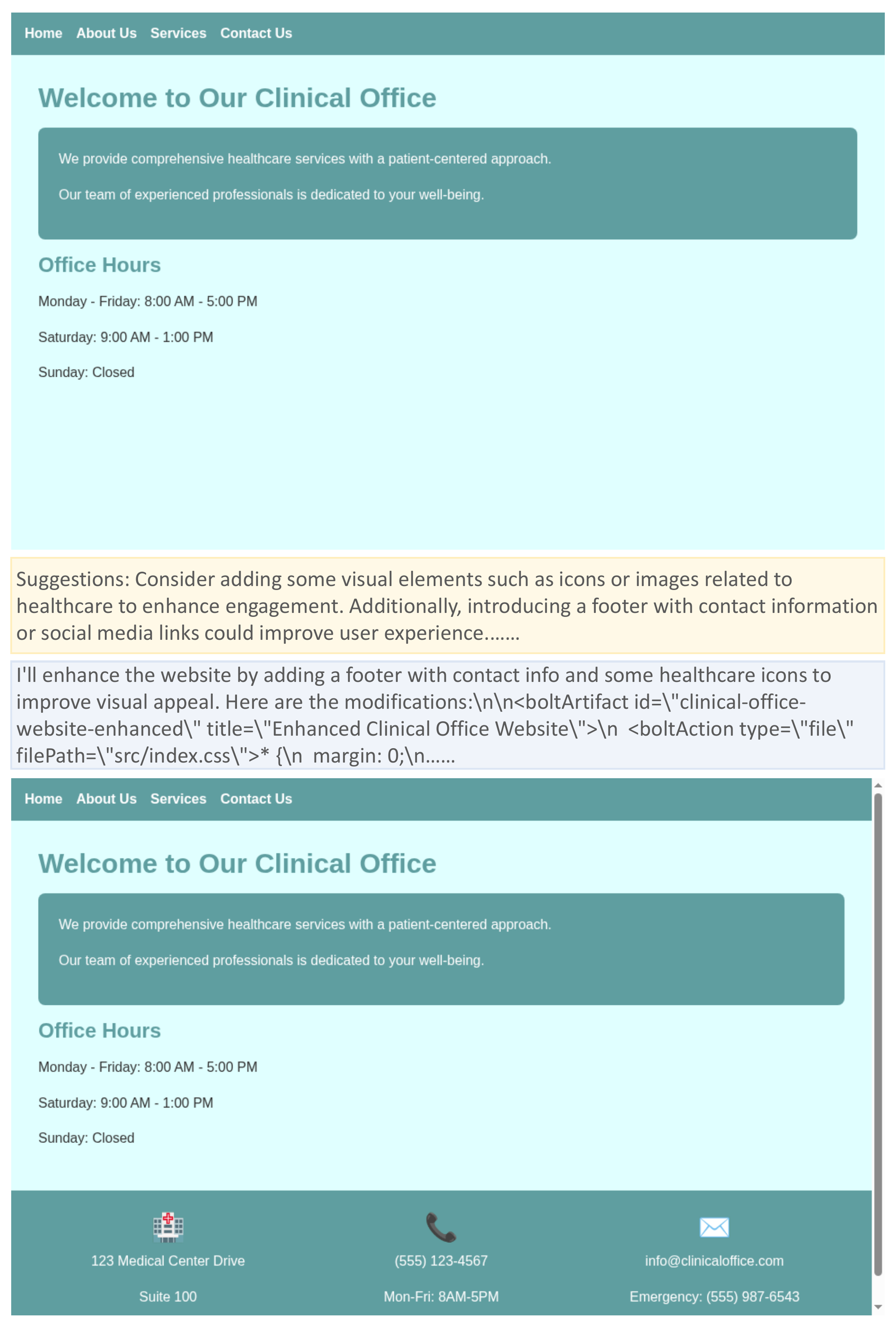}
    \caption{Example of a step in a WebGen-Agent trajectory where the agent improves the website's appearance based on screenshot feedback. The step is simplified due to space constraints.}
    
\label{fig:screenshot_improve_3}
\end{figure*}

\begin{figure*}[t]
    \centering
    \includegraphics[width=1.0\textwidth]{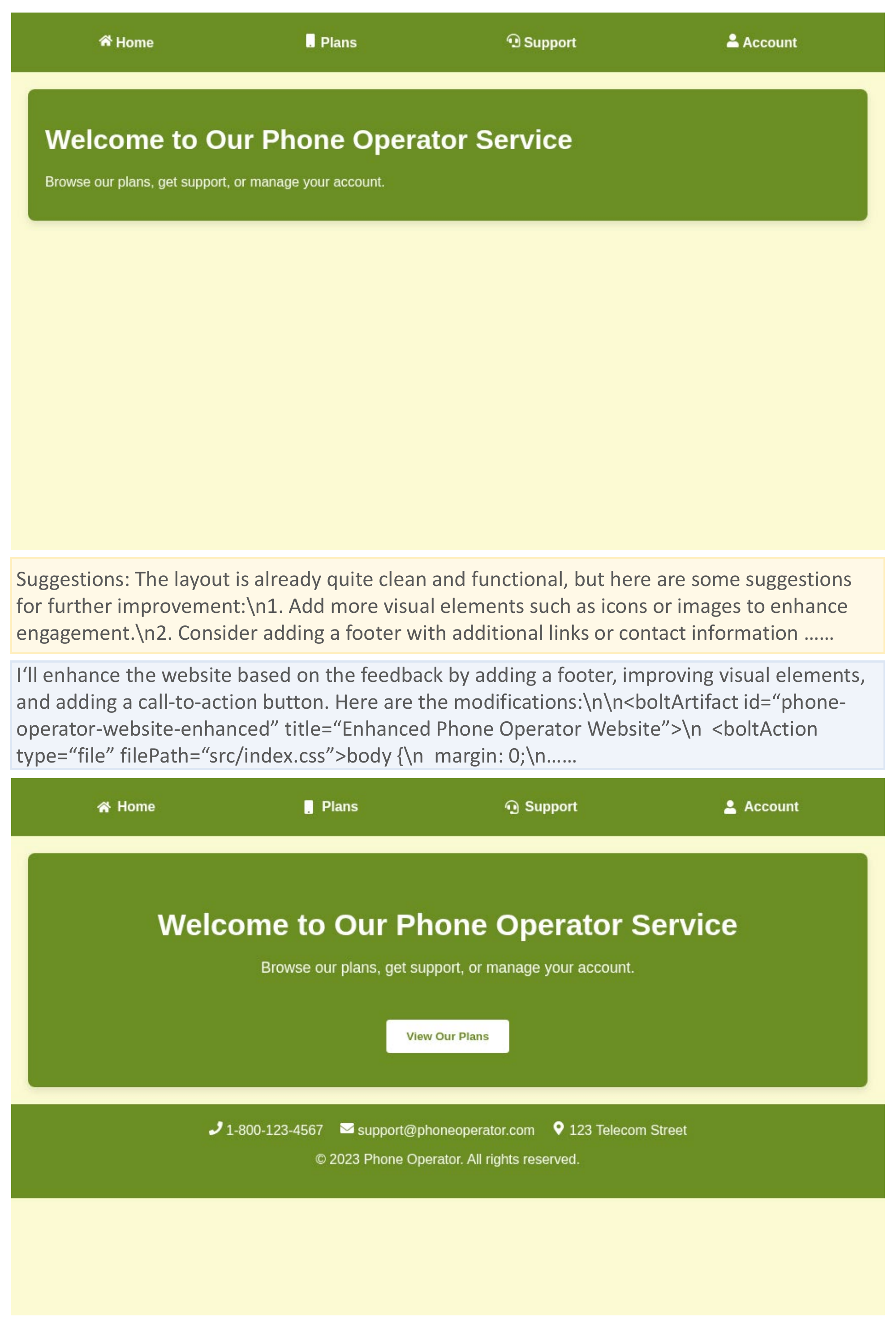}
    \caption{Example of a step in a WebGen-Agent trajectory where the agent improves the website's appearance based on screenshot feedback. The step is simplified due to space constraints.}
    
\label{fig:screenshot_improve_4}
\end{figure*}

\begin{figure*}[t]
    \centering
    \includegraphics[width=1.0\textwidth]{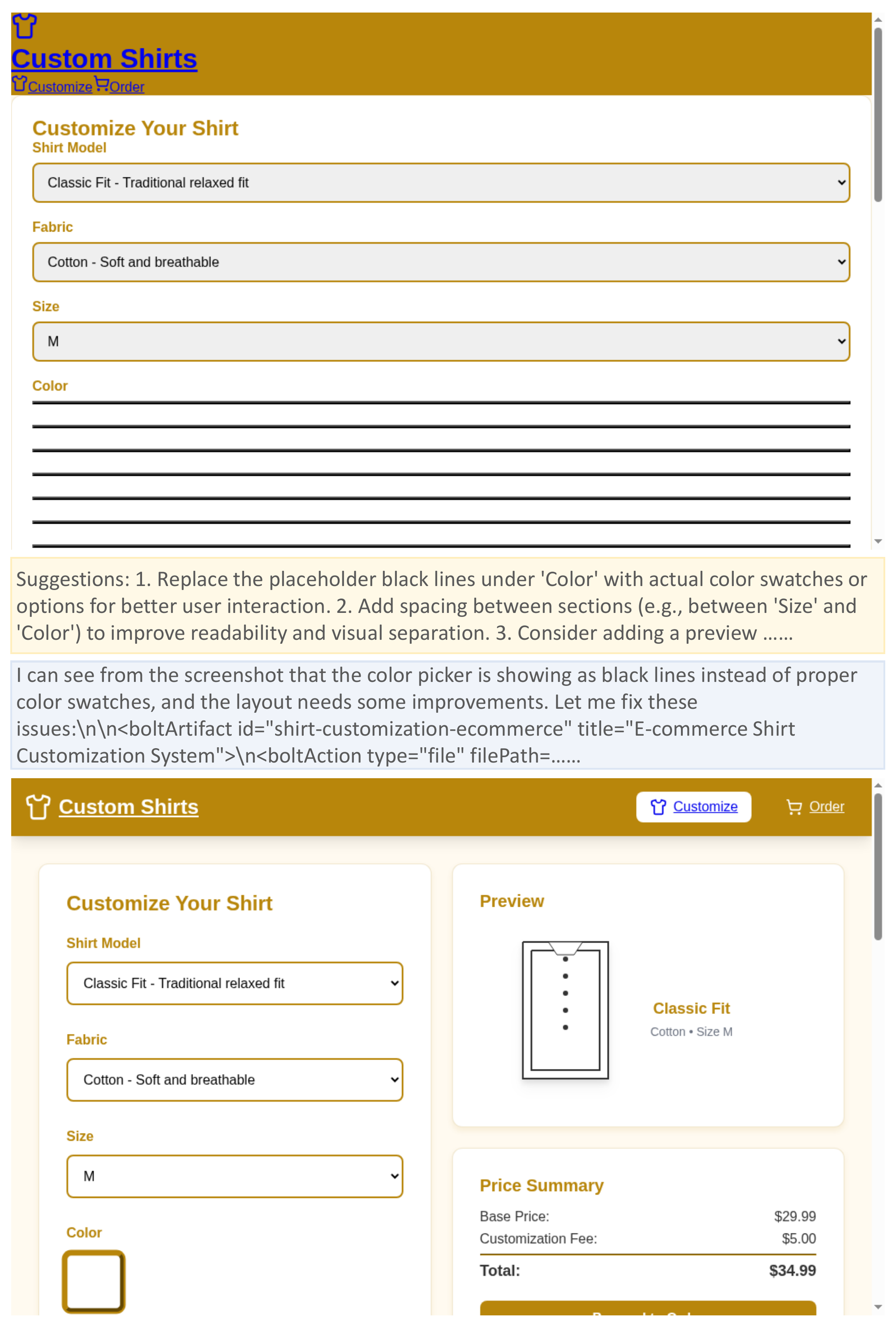}
    \caption{Example of a step in a WebGen-Agent trajectory where the agent improves the website's appearance based on screenshot feedback. The step is simplified due to space constraints.}
    
\label{fig:screenshot_improve_5}
\end{figure*}

\begin{figure*}[t]
    \centering
    \includegraphics[width=1.0\textwidth]{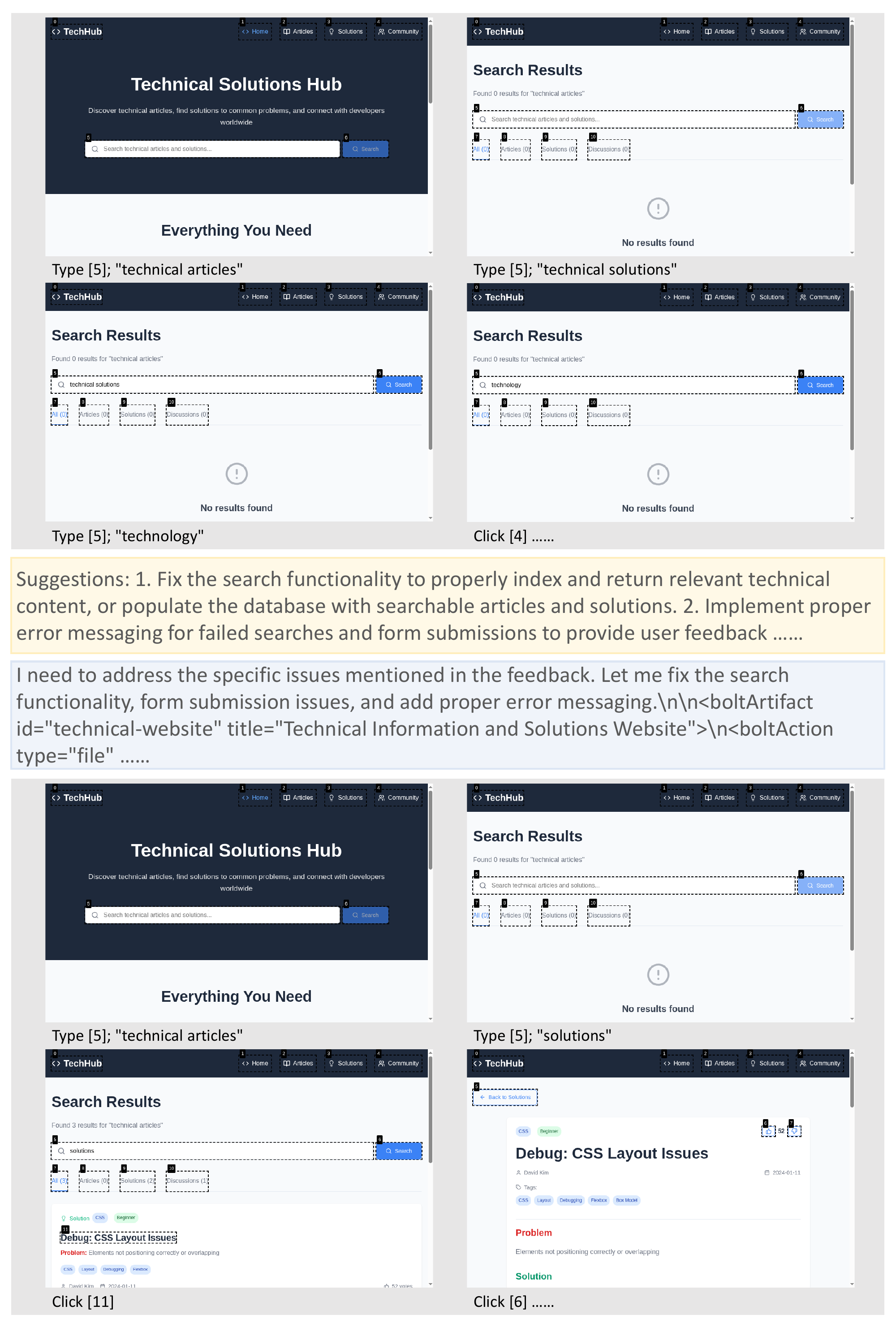}
    \caption{Example of a step in a WebGen-Agent trajectory where the agent improves the website's appearance based on GUI-agent testing feedback. The step is simplified due to space constraints.}
    
\label{fig:gui_improve_1}
\end{figure*}

\begin{figure*}[t]
    \centering
    \includegraphics[width=1.0\textwidth]{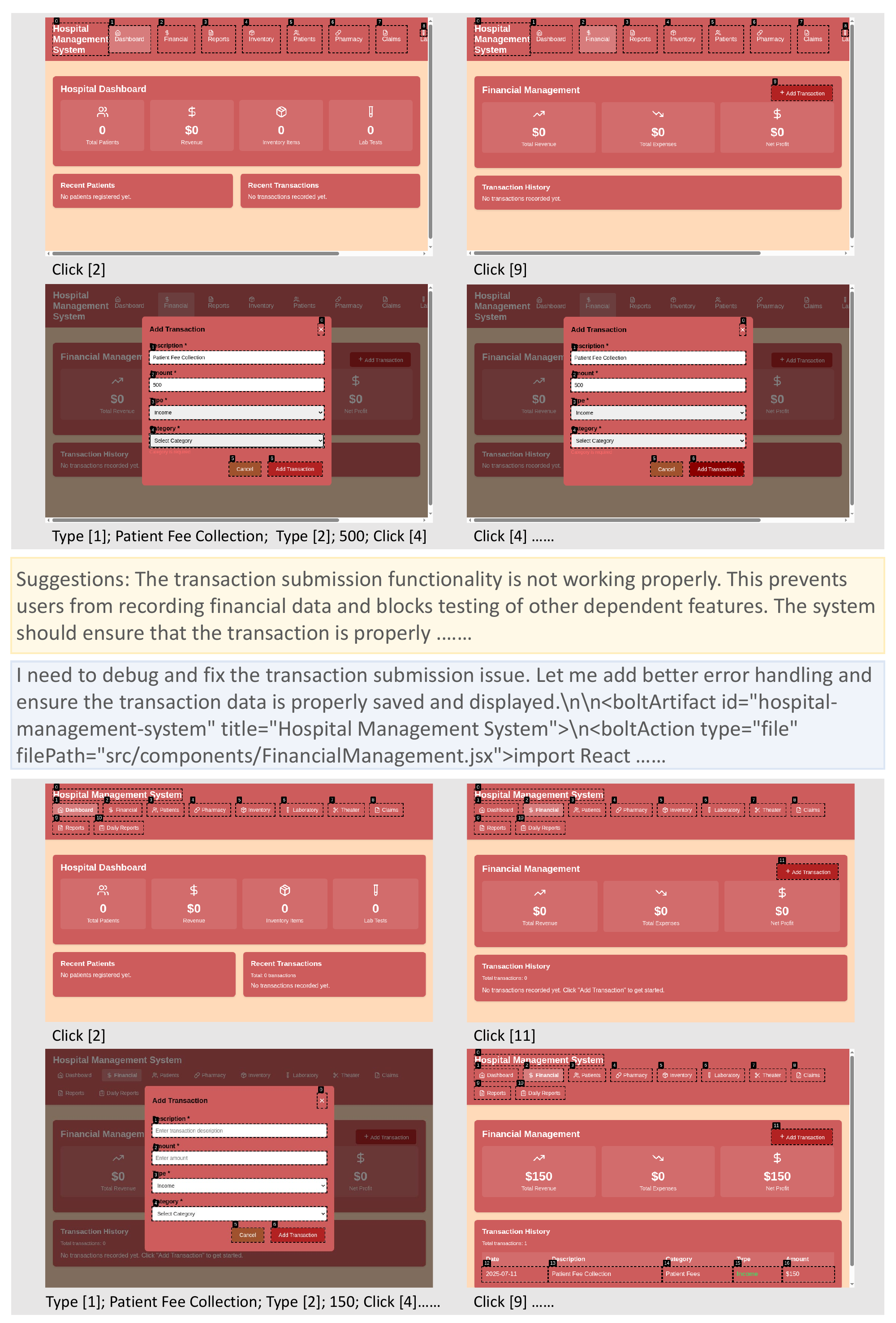}
    \caption{Example of a step in a WebGen-Agent trajectory where the agent improves the website's appearance based on GUI-agent testing feedback. The step is simplified due to space constraints.}
    
\label{fig:gui_improve_2}
\end{figure*}

\begin{figure*}[t]
    \centering
    \includegraphics[width=1.0\textwidth]{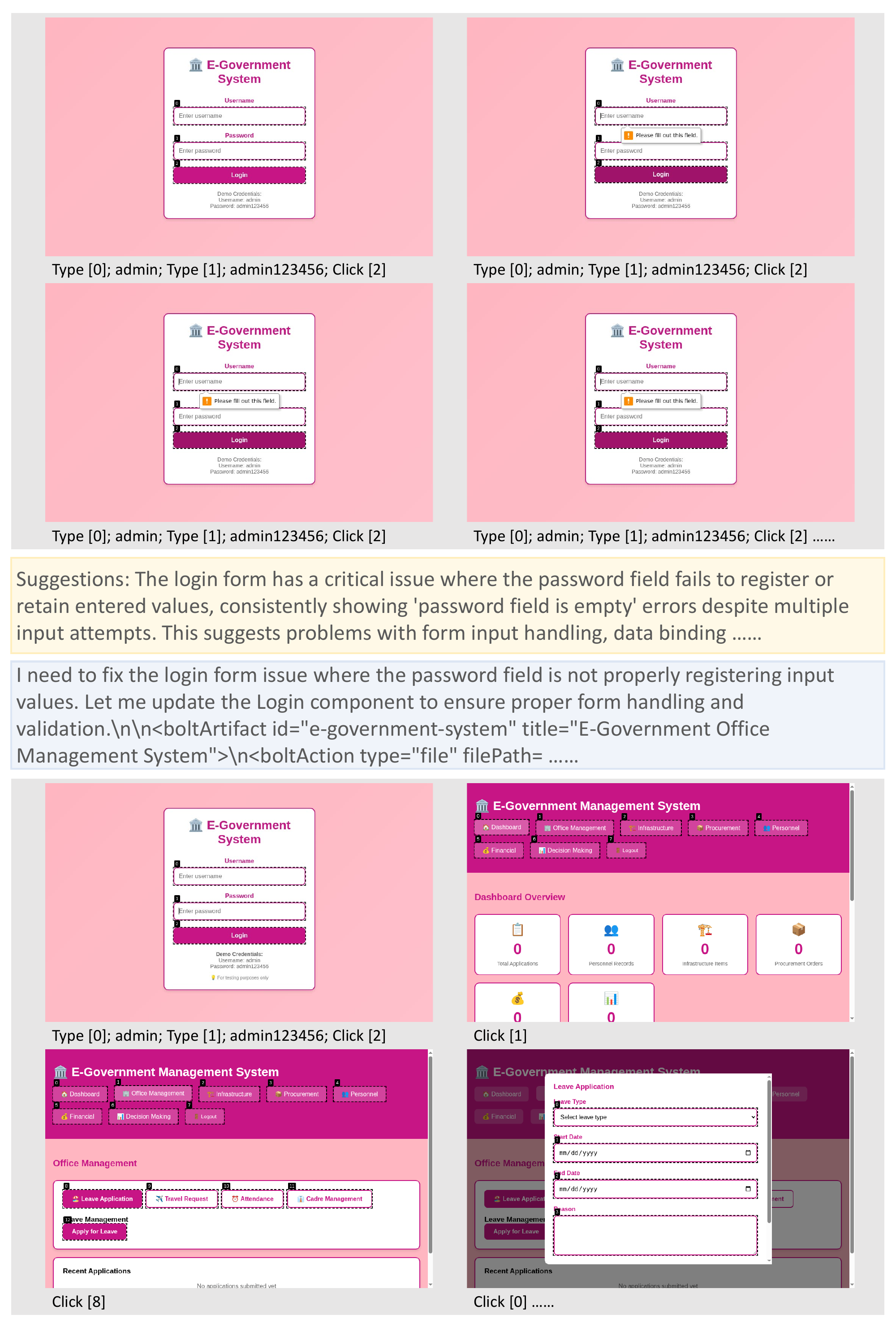}
    \caption{Example of a step in a WebGen-Agent trajectory where the agent improves the website's appearance based on GUI-agent testing feedback. The step is simplified due to space constraints.}
    
\label{fig:gui_improve_3}
\end{figure*}

\begin{figure*}[t]
    \centering
    \includegraphics[width=1.0\textwidth]{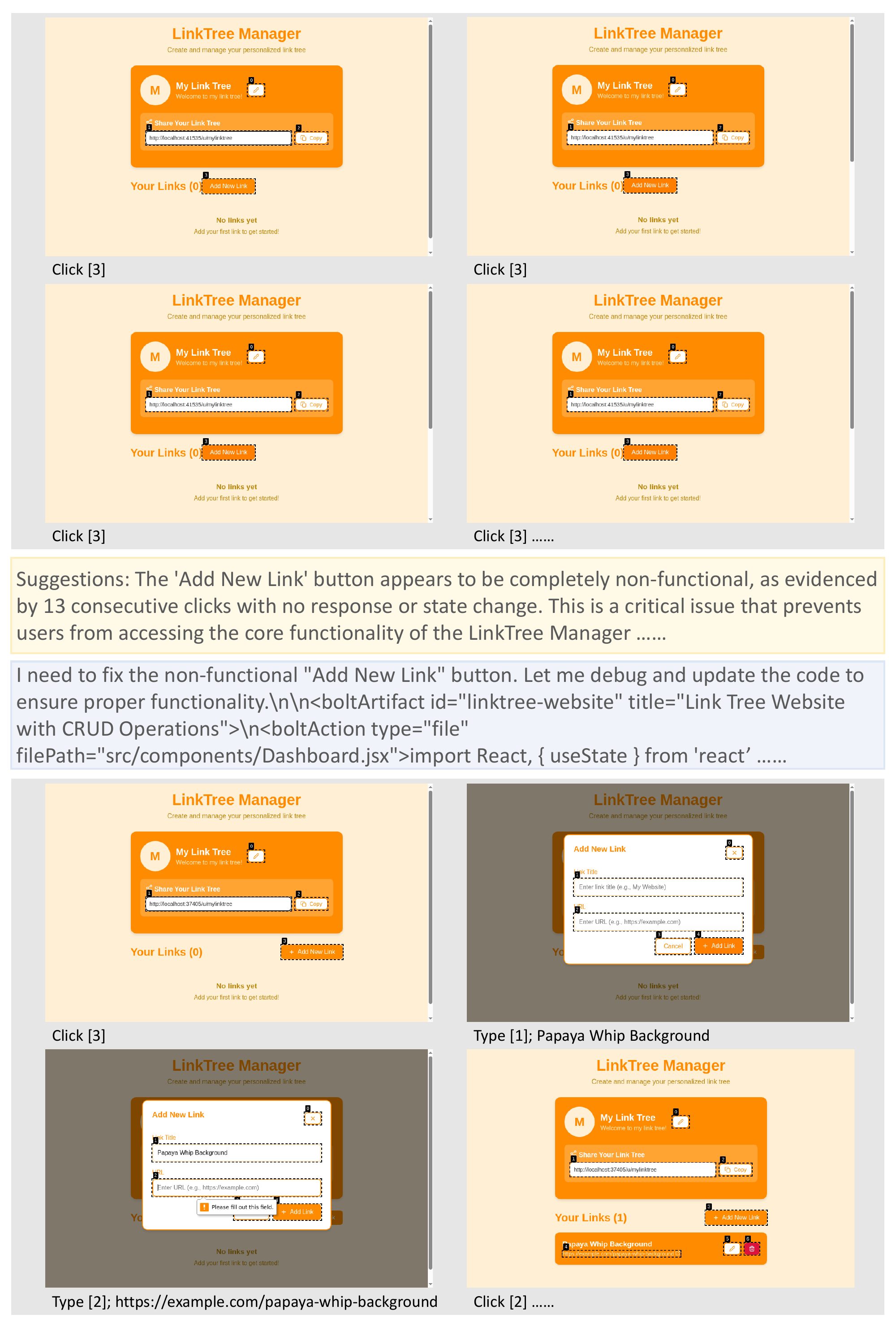}
    \caption{Example of a step in a WebGen-Agent trajectory where the agent improves the website's appearance based on GUI-agent testing feedback. The step is simplified due to space constraints.}
    
\label{fig:gui_improve_4}
\end{figure*}

\begin{figure*}[t]
    \centering
    \includegraphics[width=1.0\textwidth]{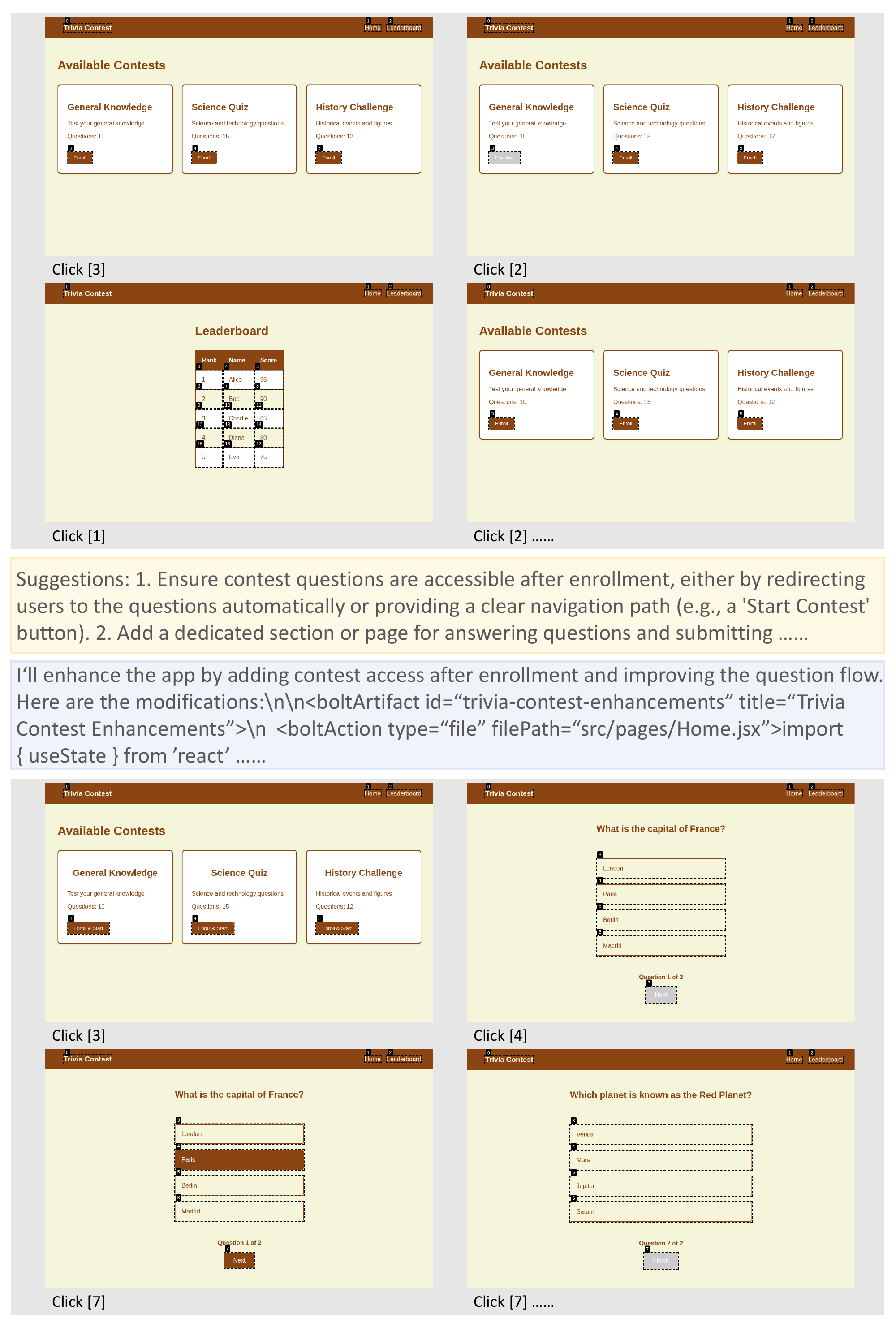}
    \caption{Example of a step in a WebGen-Agent trajectory where the agent improves the website's appearance based on GUI-agent testing feedback. The step is simplified due to space constraints.}
    
\label{fig:gui_improve_5}
\end{figure*}

To demonstrate how the WebGen-Agent workflow functions, we provide examples of steps in WebGen-Agent trajectories where the agent improves the website's appearance based on screenshot or GUI-agent feedback. As shown in Fig.~\ref{fig:screenshot_improve_1}, Fig.~\ref{fig:screenshot_improve_2}, Fig.~\ref{fig:screenshot_improve_3}, Fig.~\ref{fig:screenshot_improve_4}, and Fig.~\ref{fig:screenshot_improve_5}, the agent enhances the website's visual appeal by incorporating suggested improvements. Similarly, Fig.~\ref{fig:gui_improve_1}, Fig.~\ref{fig:gui_improve_2}, Fig.~\ref{fig:gui_improve_3}, Fig.~\ref{fig:gui_improve_4}, and Fig.~\ref{fig:gui_improve_5} illustrates how the agent refines the website's functionality based on feedback from the GUI-agent testing process. The steps are simplified due to space constraints.

\section{Usage of Large Language Models in Paper Writing}

The paper is primarily human-written. However, large language models such as o3~\citep{openai2025o3} and DeepSeek-V3~\citep{liu2024deepseek} are used to check for grammar and spelling mistakes. The words and phrases are occasionally polished by LLMs to make the wording more fluent.

\end{document}